\documentclass[lettersize,journal]{IEEEtran}

\usepackage{cite}
\usepackage{amsmath,amssymb,amsfonts}
\usepackage{algorithmic}
\usepackage{algorithm}
\usepackage{array}
\usepackage{textcomp}
\usepackage{stfloats}
\usepackage{url}
\usepackage{amssymb}
\usepackage{verbatim}
\usepackage{graphicx}
\usepackage{multirow}
\usepackage{colortbl}
\usepackage{tabularx}
\usepackage{xcolor}
\usepackage{booktabs} 
\usepackage{newfloat}
\usepackage{listings}
\usepackage{balance}
\usepackage[caption=false,font=footnotesize,labelfont=rm,textfont=rm]{subfig}

\usepackage[pagebackref,breaklinks,colorlinks]{hyperref}
\hypersetup{
	citecolor=blue,
	linkcolor=red,
	urlcolor=magenta,
}
\definecolor{mygreen}{RGB}{34,139,34} 
\def\BibTeX{{\rm B\kern-.05em{\sc i\kern-.025em b}\kern-.08em
		T\kern-.1667em\lower.7ex\hbox{E}\kern-.125emX}}
\begin{document}
\title{Fourier Series Coder: A Novel Perspective on Angle Boundary Discontinuity Problem for Oriented Object Detection}
\author{
	\IEEEauthorblockN{
		Minghong Wei$^\dagger$, Pu Cao, Zhihao Chen, Zhiyuan Zang, Lu Yang$^*$ and Qing Song
	}
	\thanks{Corresponding Author: Lu Yang.
		
	\hspace{0.1em}Minghong Wei, Pu Cao, Zhihao Chen, Zhiyuan Zang, Lu Yang, and Qing Song are with the School of Intelligent Engineering and Automation, Beijing University of Posts and Telecommunications, Beijing, 100082,
	China (e-mail: \{wmh0517, caopu, zhihaochen666, 2025010590, soeaver, priv\}@bupt.edu.cn).}
}

\maketitle

\begin{abstract}
With the rapid advancement of intelligent driving and remote sensing, oriented object detection has gained widespread attention. However, achieving high-precision performance is fundamentally constrained by the Angle Boundary Discontinuity (ABD) and Cyclic Ambiguity (CA) problems, which typically cause significant angle fluctuations near periodic boundaries. Although recent studies propose continuous angle coders to alleviate these issues, our theoretical and empirical analyses reveal that state-of-the-art methods still suffer from substantial cyclic errors. We attribute this instability to the structural noise amplification within their non-orthogonal decoding mechanisms. This mathematical vulnerability significantly exacerbates angular deviations, particularly for square-like objects. To resolve this fundamentally, we propose the Fourier Series Coder (FSC), a lightweight plug-and-play component that establishes a continuous, reversible, and mathematically robust angle encoding-decoding paradigm. By rigorously mapping angles onto a minimal orthogonal Fourier basis and explicitly enforcing a geometric manifold constraint, FSC effectively prevents feature modulus collapse. This structurally stabilized representation ensures highly robust phase unwrapping, intrinsically eliminating the need for heuristic truncations while achieving strict boundary continuity and superior noise immunity. Extensive experiments across three large-scale datasets demonstrate that FSC achieves highly competitive overall performance, yielding substantial improvements in high-precision detection. The code will be available at \url{https://github.com/weiminghong/FSC}.
\end{abstract}

\section{Introduction}

Oriented object detection has emerged as a fundamental yet highly challenging task in computer vision, serving as an indispensable technique for accurately capturing object poses in densely packed and complex scenarios. 
Unlike standard object detection that relies on axis-aligned Horizontal Bounding Boxes (HBBs), oriented detection employs Oriented Bounding Boxes (OBBs) to provide rigorous geometric grounding and fine-grained spatial perception. This robust representation is critical for a broad spectrum of real-world applications, including remote sensing imagery \cite{ding2019learning,han2021align,yang2021r3det,ming2021cfc,yu2025wholly}, scene text recognition \cite{wang2025s3inet,liao2018rotation,ma2018arbitrary}, intelligent retail \cite{hu2024efficient}, and robotic grasping \cite{chu2018real, kumra2020antipodal}

\begin{figure}[t]
	\centering{\includegraphics[width=\columnwidth]{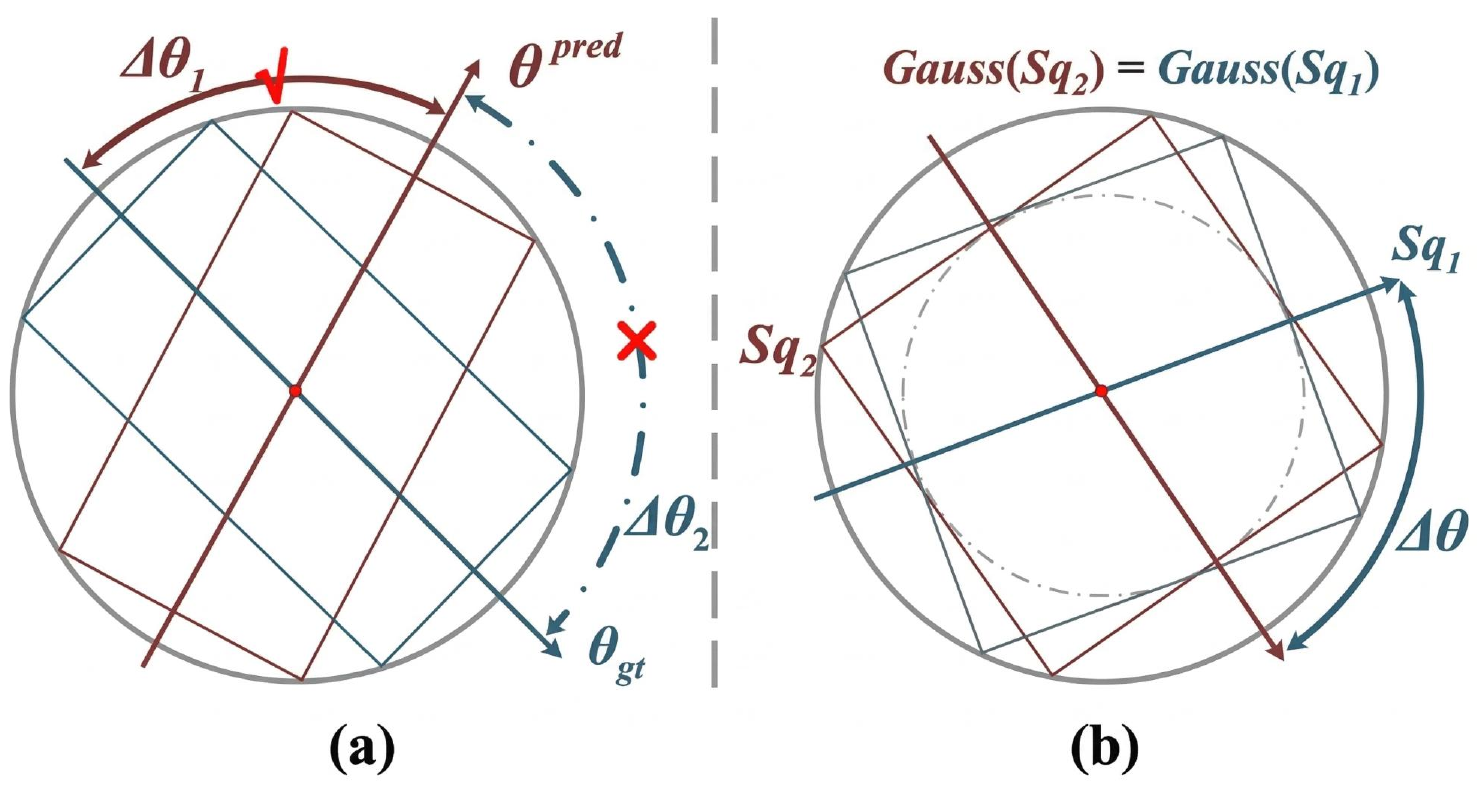}}
	\caption{Examples of discontinuities caused by rotational symmetry.
		(a) Angle Boundary Discontinuity: Detectors cannot distinguish the optimal path ($\Delta {\theta _1}$ vs. $\Delta {\theta _2}$).
		(b) Cyclic Ambiguity: The square OBBs with different orientations can correspond to the same Gaussian distribution.}
	\label{title1}
\end{figure}

However, a persistent bottleneck in optimizing OBBs $\mathcal{B}(x, y, w, h, \theta)$ is the \textbf{Angle Boundary Discontinuity} (\textbf{ABD}) problem \cite{yang2020arbitrary,qian2021learning,yang2019scrdet,nie2022multi}. While HBBs allow for smooth, continuous regression across spatial parameters, the angular periodicity in OBBs introduces severe numerical instability. As illustrated in Fig. \ref{title1}\textcolor{red}{(a)}, detectors struggle to differentiate the optimal shortest rotation path ($\Delta\theta_1$) from the suboptimal longer path ($\Delta\theta_2$). The latter forces the network to compensate through complex, simultaneous adjustments of coupled parameters (e.g., length and width), leading to a highly unstable loss landscape. Early attempts to mitigate these sharp loss surges via smoothed loss functions \cite{qian2021learning,yang2019scrdet} yielded marginal improvements.

\begin{figure*}[t]
	\centering
	
	\subfloat[Rectangular object.]{
		\label{title21}
		\includegraphics[width=0.235\linewidth]{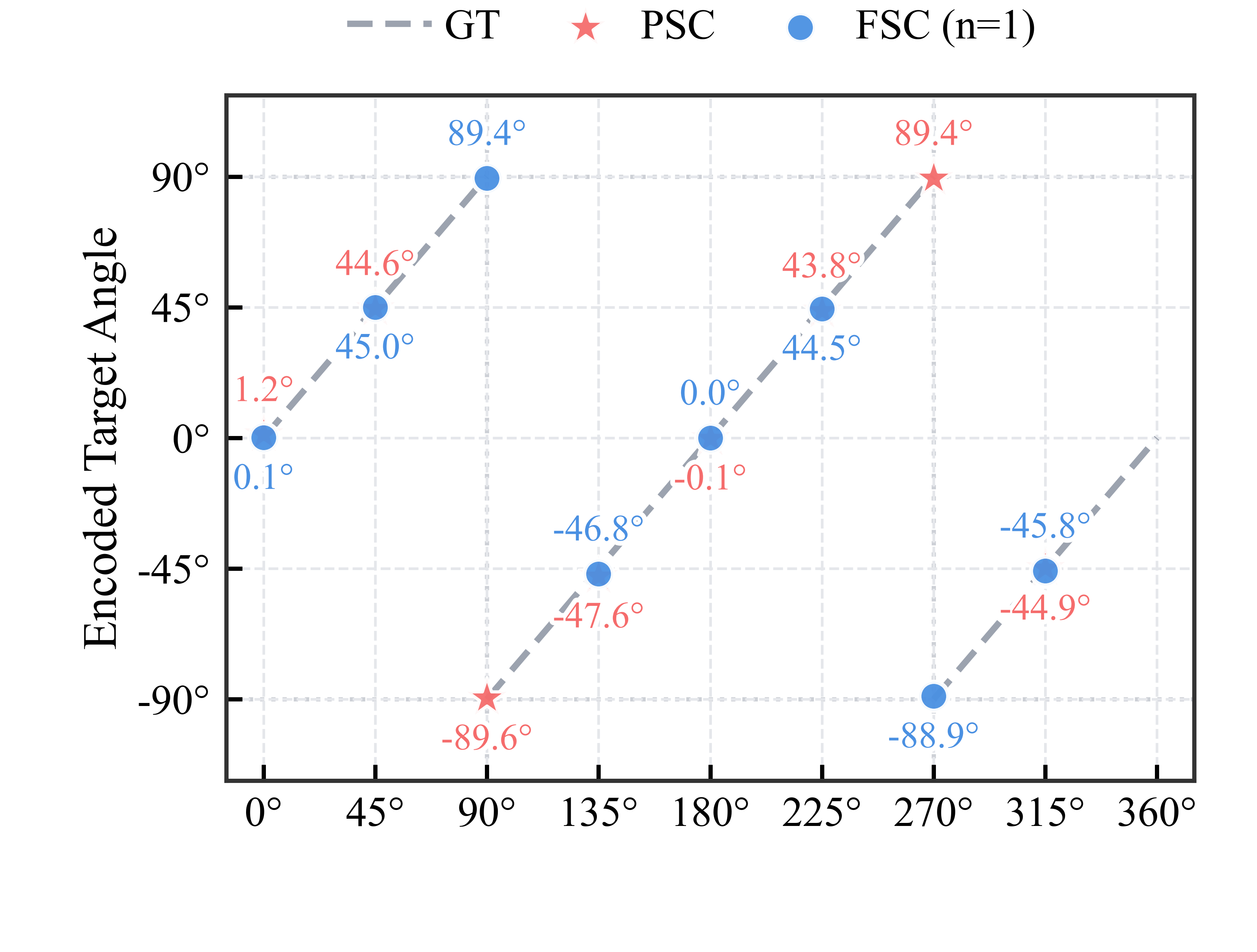}
	}
	\hfil
	\subfloat[Square-like object.]{
		\label{title22}
		\includegraphics[width=0.235\linewidth]{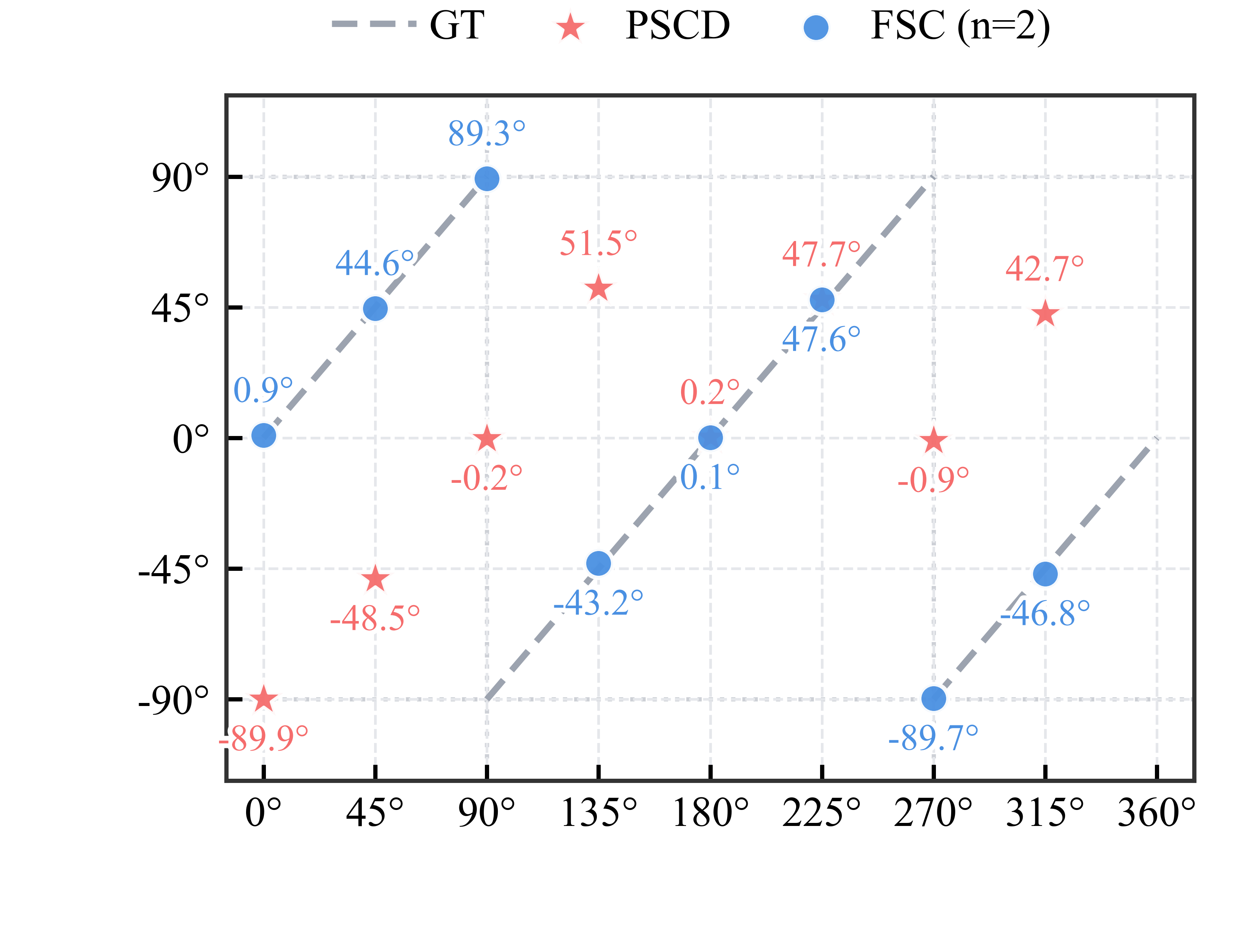}
	}
	\hfil
	\subfloat[Simulated Modulus Collapse.]{
		\label{title23}
		\includegraphics[width=0.235\linewidth]{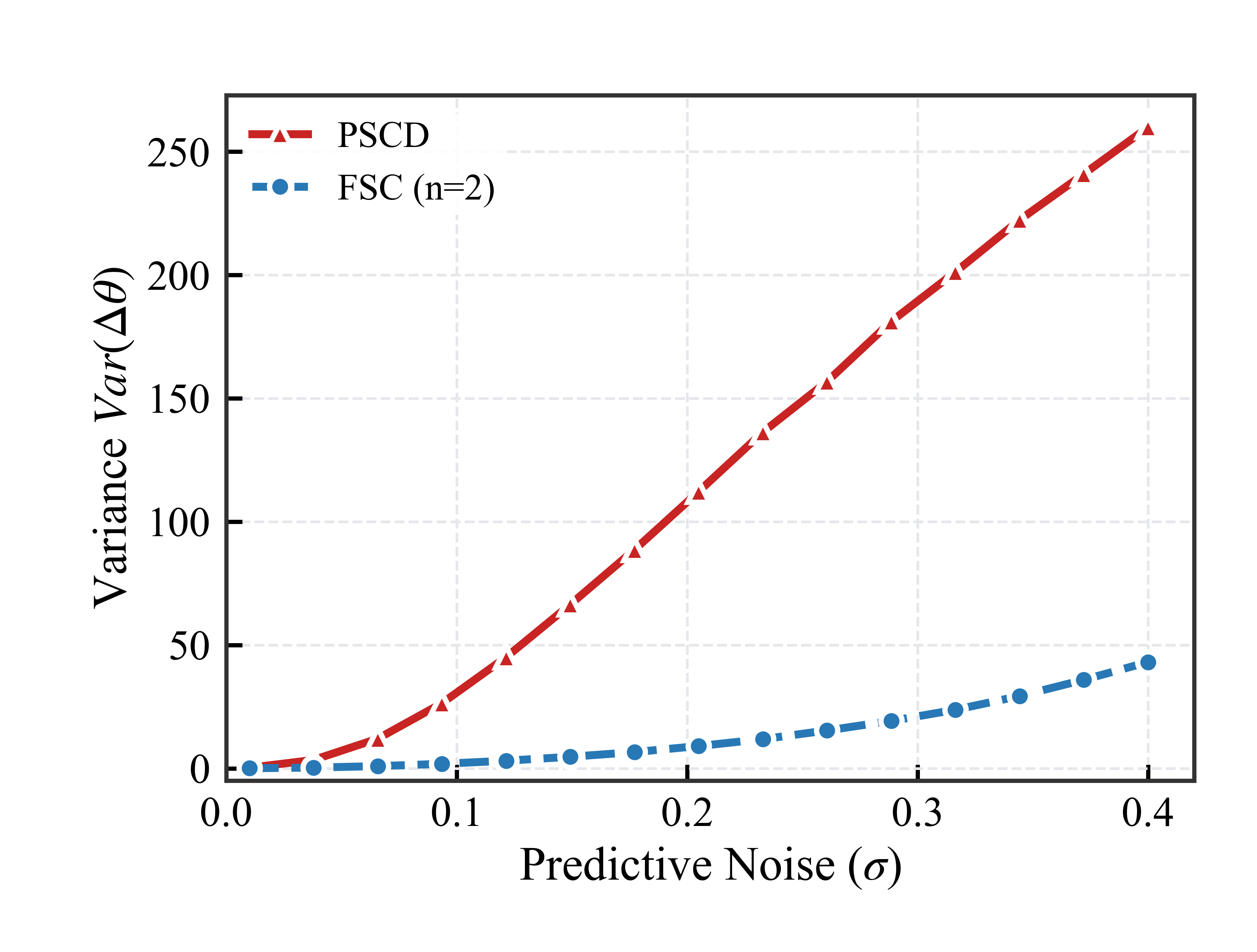}
	}
	\hfil
	\subfloat[Decoding Error Distribution.]{
		\label{title24}
		\includegraphics[width=0.235\linewidth]{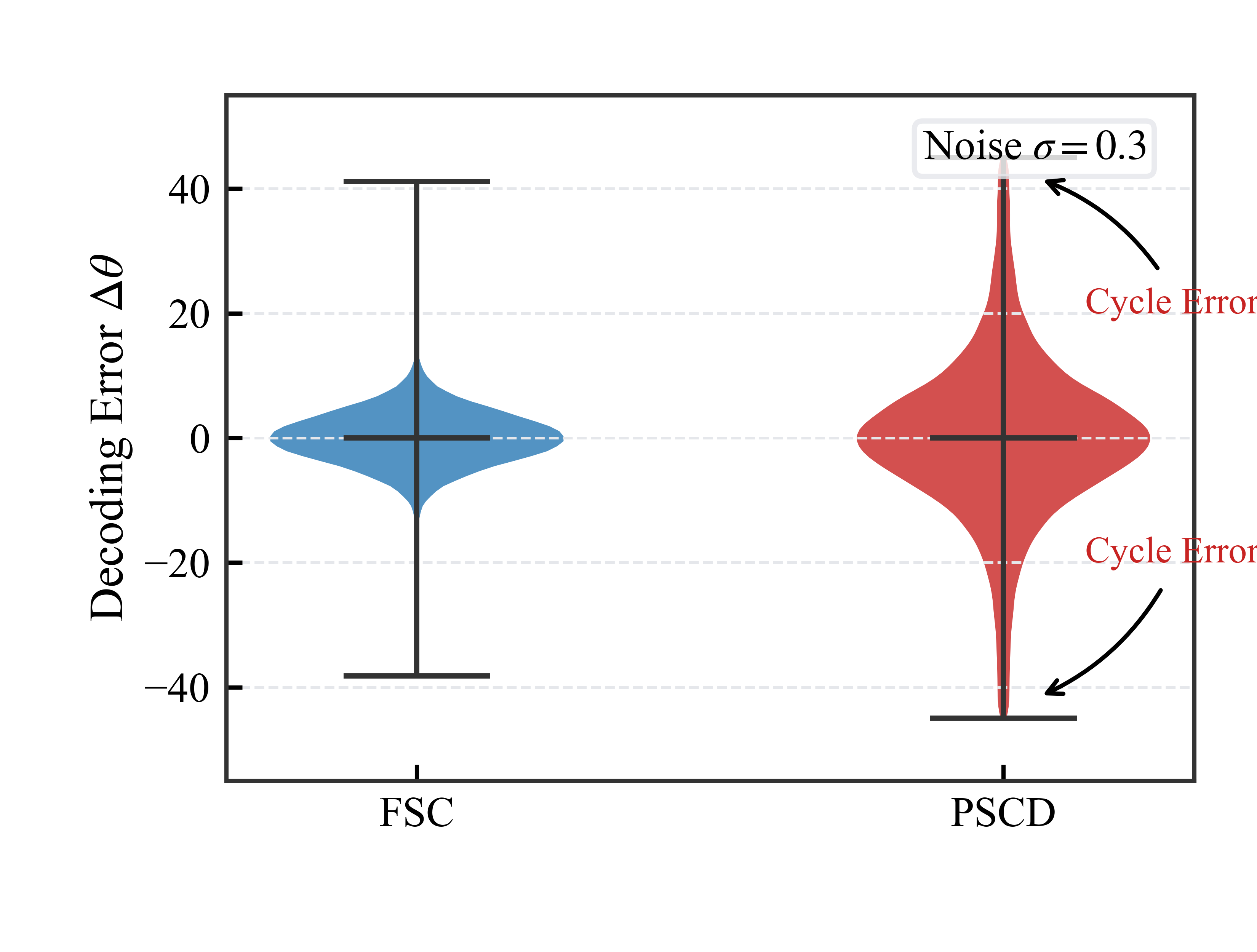}
	}
	\caption{Comprehensive comparison of decoding stability under predictive noise. (a) and (b) present the discrete angle predictions for a rectangular and a square-like object, respectively. (c) Monte Carlo simulation of decoding variance under simulated modulus collapse. Without explicit geometric constraints, the non-orthogonal PSCD \cite{yu2023phase} suffers from an exponential explosion in variance. (d) Error distribution at high noise ($\sigma=0.3$). The unconstrained PSCD exhibits severe cycle errors, whereas our manifold-constrained FSC maintains a highly compact and robust decoding manifold.}
	\label{pic2}
\end{figure*}

To circumvent direct angular regression, a prominent line of research shifted towards implicit joint optimization. Mainstream OBB coders \cite{yang2021rethinking,yang2021learning,yang2022kfiou,marques2025gaucho,dinggsdet} transform bounding boxes into Gaussian distributions (e.g., KLD \cite{yang2021learning}). While these methods achieve significant performance gains, they are fundamentally constrained by the \textbf{Cyclic Ambiguity (CA)} problem, which stems from the encoding process itself. As shown in Fig. \ref{title1}\textcolor{red}{(b)}, distinct physical OBBs can be encoded into mathematically identical Gaussian distributions, confusing the network during decoding. Similarly, proposal-driven methods \cite{qiao2023novel,xu2020gliding,nie2022multi,xiao2024theoretically,zhu2020adaptive,zhou2020arbitrary, xu2023gaussian} design complex rotated anchors to implicitly optimize orientation. Ultimately, these methods effectively sidestep the ABD problem rather than resolving it fundamentally.

Recognizing the limitations of implicit modeling, a more direct paradigm has emerged: explicit angle coders. Early explorations, such as CSL-based frameworks \cite{yang2020arbitrary,yang2021dense,zeng2023ars,wang2022multigrained}, framed angle prediction as a discrete classification task, but their high-precision potential is severely limited by quantization errors. To achieve a continuous representation, state-of-the-art methods like PSC \cite{yu2023phase} map angles into a continuous optical phase space. 

Although PSC elegantly addresses ABD and CA in theory, we reveal that its unconstrained overcomplete design is topologically vulnerable. To empirically validate this, we conduct a fine-grained continuous rotational analysis across a full $360^\circ$ span. Specifically, we generate a sequence of images by rotating a single image through $360^\circ$ in $1^\circ$ increments, and then sequentially input these images into the same detector for inference. As illustrated in Fig. \ref{pic2}\textcolor{red}{(a)}, while both PSC and our proposed method successfully fit the angular variations of rectangular objects, the non-orthogonal three-phase basis utilized in PSCD inherently amplifies underlying neural predictive noise. Consequently, unexpected cycle errors emerge for square-like objects, as shown in Fig. \ref{pic2}\textcolor{red}{(b)}. 
The Monte Carlo simulation presented in Fig. \ref{pic2}\textcolor{red}{(c)} reveals the root cause, demonstrating that this unconstrained Cartesian synthesis makes the encoded features highly susceptible to modulus degradation during optimization. Once the feature modulus drops, the decoding variance increases exponentially. Under noticeable noise perturbations, the diminished Cartesian vectors frequently cross the origin, triggering $\arctan2$ singularities that directly manifest as the significant phase-wrapping outliers observed in Fig. \ref{pic2}\textcolor{red}{(d)}.
While heuristic hard thresholds (e.g., overriding the angle to $0^\circ$ when \textit{phase\_mod} $< 0.47$) are often employed to bypass these chaotic regions, they inevitably reintroduce discontinuous representations, underscoring the critical need for an intrinsically stable encoding strategy.

Recognizing that an optimal encoding strategy must intrinsically guarantee mathematical continuity, reversibility, and orthogonal stability, we propose the Fourier Series Coder (FSC). Inspired by the continuous phase representation but discarding the flawed overcomplete design, FSC fundamentally restructures the encoding paradigm. Specifically, rather than relying on a non-orthogonal basis that induces false truncations during high-frequency phase unwrapping, FSC strictly maps the target angle $\theta$ to a minimal orthogonal Fourier series basis $fsc(\cdot)$. This foundational design inherently eliminates structural redundancy and mitigates the mathematical amplification of predictive noise.

\begin{figure*}[t]
	\centering{\includegraphics[width=2.0\columnwidth]{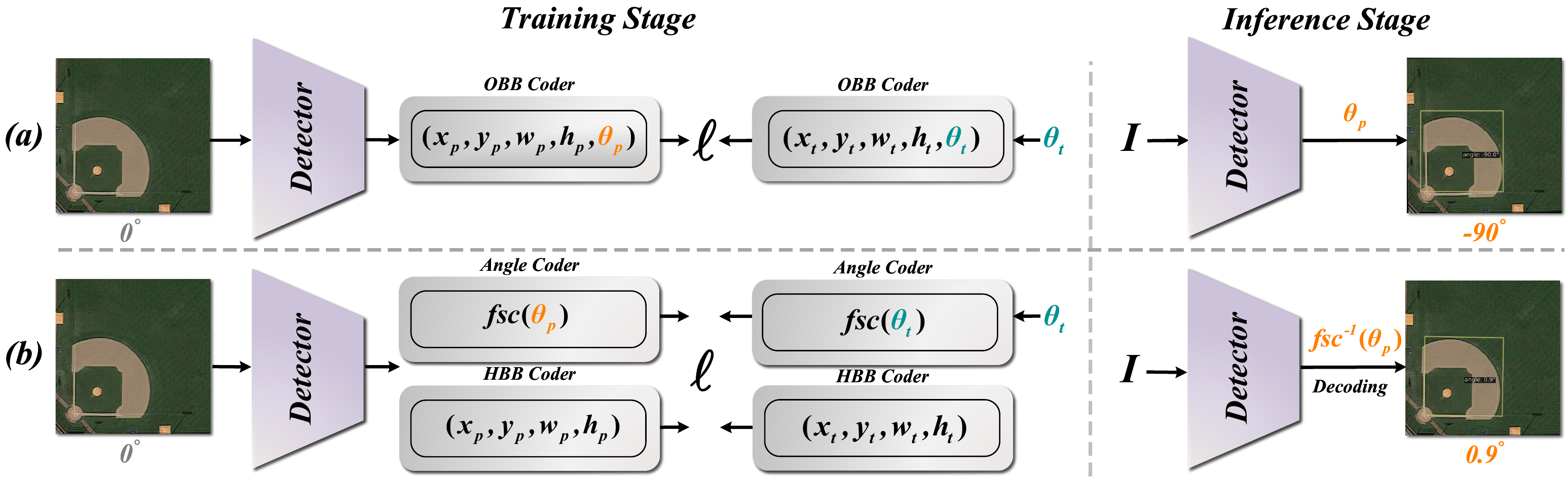}}
	\caption{The optimization paradigm of the proposed Fourier Series Coder. 
		(a) The OBB coder encodes the angle with other parameters together and uses joint optimization functions.  (b) FSC encodes the angle as continuous components, and its optimization is decoupled from other parameters. The testing stage determines the predicted angle $\theta_p$ through decoding Fourier series components using $fsc^{-1}(\cdot)$.
	}
	\label{main}
\end{figure*}

Crucially, to proactively prevent the feature modulus collapse that degrades previous unconstrained decoders, we explicitly introduce a geometric manifold constraint alongside the orthogonal basis. This regularization effectively suppresses structural noise inflation. As demonstrated in Fig. \ref{pic2}\textcolor{red}{(c)}, FSC successfully preserves the integrity of the continuous decoding manifold $fsc^{-1}(\cdot)$ across the entire $360^\circ$ space. By relying entirely on this strict mathematical mapping, FSC achieves highly stable dual-frequency unwrapping without resorting to fragile heuristic thresholds, effectively resolving both ABD and CA problems. Furthermore, as illustrated in Fig. \ref{main}, this reversible encoding design enables the inverse function $fsc^{-1}(\cdot)$ to output robust angle predictions $\theta_p$ during inference. Finally, a decoupling paradigm is employed to prevent the negative coupling associated with directly optimizing Oriented Bounding Boxes (OBBs). Specifically, our contributions can be summarized as follows:

1) We systematically analyze the fundamental limitations of existing continuous angle coders, identifying for the first time how unconstrained non-orthogonal designs lead to feature modulus collapse, ultimately resulting in severe cyclic errors.

2) We propose the Fourier Series Coder (FSC), which fundamentally resolves both ABD and CA problems by guaranteeing decoding stability through a minimal orthogonal Fourier basis and an explicit geometric manifold constraint.

3) Extensive experimental results validate the significant competitiveness and robustness of FSC.

\section{Related Work}

In oriented object detection tasks, boundary discontinuity has always been a challenging technical problem. Understanding the strategies (e.g., encoding function $f$ and loss function $\ell$) adopted by existing research and their potential impact is crucial to grasping the core contributions of this paper. The proposed solutions to the series of problems caused by discontinuous representations can be summarized into three categories: smoothing loss function, OBB coder, and angle coder.  
\subsection{Loss Attempt}

Several strategies initially aimed to alleviate the ABD problem through joint-optimized loss functions $\ell(x,y,w,h,\theta)$ with geometric parameters.
PIoU \cite{chen2020piou} minimizes the angular deviation via Peixels-IoU. RSDet \cite{qian2021learning} eliminates loss discontinuity by stabilizing Rotation Sensitive Error (RSE). SCRDet \cite{yang2019scrdet} controls the magnitude and direction of gradient descent through optimization factors such as IoU, thereby avoiding abrupt loss changes. While these initial attempts provide limited relief from ABD, they do not fundamentally address the core issue of continuous encoding.
\subsection{Oriented Bounding Box Coder}

Gaussian-based methods (e.g., GWD \cite{yang2021rethinking}, KLD \cite{yang2021learning}, and KFIoU \cite{yang2022kfiou}) transform OBBs into Gaussian distributions $Gauss(x,y,w,h,\theta)$ and implicitly optimize them using matched loss functions (e.g., $\ell_{gwd}$ and $\ell_{kld}$) to circumvent direct angular regression.
GWD \cite{yang2021rethinking} asserts that rotated IoU is non-differentiable and establishes a joint optimization loss between Gaussian distributions via the Gaussian-Wasserstein distance. KLD \cite{yang2021learning} refines GWD by dynamically adjusting parameter gradients based on object characteristics, addressing scale invariance but not resolving the square-like object issue. KFIoU \cite{yang2022kfiou}, on the other hand, approximates the IoU of skewed bounding boxes (IoUSkew) through Kalman Filtering IoU (KFIoU). Although these similar joint Gaussian-based optimization methods effectively alleviate the discontinuity problem, the potential cyclic ambiguity remains. ProbIoU \cite{murrugarra2024probabilistic} proposes a probabilistic IoU metric derived from Gaussian representations to further enhance the alignment and evaluation of oriented bounding boxes.	  
Other OBB coders, such as those discussed in \cite{qiao2023novel,xu2020gliding,nie2022multi}, have been proposed to design complex proposals $p(x,y,w,h,\theta)$ within two-stage models, yet they still rely on implicit optimization.

\subsection{Angle Coder}

CSL \cite{yang2020arbitrary} highlights that encoding discontinuity primarily arises from the Periodicity of Angle (PoA) and the Exchangeability of Edges (EoE). CSL formulates angle prediction as a classification task. Specifically, its encoding function $f=onehot(\theta)$, and optimized by classification loss $\ell_{focal}$.
However, the discrete encoding affects high-precision metrics, and the model suffers from an oversized classification head. 
Subsequently, CSL-based methods have been proposed to optimize sparse encoding (e.g., DCL \cite{yang2021dense}, AR-CSL \cite{zeng2023ars}, and MGAR \cite{wang2022multigrained}) or design novel classification losses (e.g., GF-CSL \cite{wang2022gaussian}). Nevertheless, none of the methods mentioned above fully satisfy the continuity of encoding.

PSC \cite{yu2023phase} marks a significant advance by introducing the multi-phase shift and cyclic wrapping to encode angles. ACM \cite{xu2024rethinking} argues that the key to resolving the boundary problem is smooth function encoding, rather than joint or independent optimization.
Similar methods \cite{10247155,10521673,xu2024rethinking} attempt to enhance detection performance by increasing the encoding length, failing to consider its potential impact. More critically, as pointed out by COBB \cite{xiao2024theoretically}, although PSC \cite{yu2023phase} and ACM \cite{xu2024rethinking} encode the angle as continuous vectors, discontinuity still exists for square-like objects. 
In summary, existing angle coders are similarly unable to eliminate Cyclic Ambiguity (CA). Our experiments demonstrate that this ambiguity is attributable to the inherent numerical fluctuation in the predicted encoding components, which consequently leads to unacceptable errors during the decoding stage.
Building upon continuous angle representation, our FSC employs a more robust encoding-decoding strategy and reversible mapping to ensure reliable prediction values.

\subsection{Others}

Some state-of-the-art frameworks aim to bypass direct angle regression entirely by proposing novel spatial transformation strategies. Notable examples include transferring angle prediction into scale regression \cite{song2024direction}, or employing hierarchical mask prompting to implicitly capture multi-oriented semantics \cite{yao2024hierarchical}. In parallel, several recent advances focus on enhancing the model's spatial understanding through frequency-domain analysis and dynamic label assignment. For instance, FAAFusion \cite{gu2026fourier} introduces an FSAA module based on Fourier rotation invariants to accurately extract the object's principal direction in the frequency domain. Building on this frequency-centric perspective, wavelet-based energy label reassignment \cite{song2025efficient} utilizes frequency-domain information to guide the model's focus toward the essential spatial distribution of rotated targets. Additionally, category-aware dynamic label assignment strategies \cite{feng2025category} have been proposed to adaptively select high-quality proposals based on the specific geometric characteristics of different categories.
While these implicit or semi-implicit representations \cite{dai2022ace, xu2025instance, dang2024det, wang2023bounding} significantly improve overall detection metrics, they often introduce complex architectural overhead or rely on parameter-sensitive assignment strategies.
In contrast, our proposed FSC mathematically resolves the boundary discontinuity at the foundational encoding level, offering a lightweight, strictly explicit, and inherently robust orientation representation.

\section{Method}
\label{Method}
As analyzed earlier, existing angle coders (such as KLD \cite{yang2021learning}, CSL \cite{yang2020arbitrary}, and PSC \cite{yu2023phase}) are either inherently limited by angle boundary discontinuities (ABD) or lack the necessary reversible and robust encoder-decoder strategies, which exacerbate Cyclic Ambiguity (CA).
To address these challenges, our Fourier Series Coder (FSC) is proposed as a continuous, reversible, and robust angle coder. 
Its design principles and implementation details in Sec. \ref{sec:fsc}.
The remainder of this section is organized as follows: Sec. \ref{sec:Immunity} analyzes its superior noise immunity over existing approaches. Finally, Sec. \ref{sec:Loss} introduces the corresponding model architecture and loss function.
\subsection{Fourier Series Coder}
\label{sec:fsc}
\textbf{Encoding.} Taking the long edge 90-degree (le90) definition, as an example, the remaining variables can be defined as follows: 
\begin{itemize}
	\item $\theta:$ Input angle, in range ${[-}\pi {/2,}\pi {/2)}$
	\item $\gamma:$ Expanded angle, in range ${[-}\pi {,}\pi {)}$
	\item $2N+1:$ The number of Fourier series terms
\end{itemize}

Since the ground truth angle intrinsically possesses a period of $2\pi$, whereas the Oriented Bounding Box (OBB) typically operates under a period of $\pi$ (e.g., $le90$ format), an angle scaling mapping is required to align the periods:
\begin{equation}
	\gamma = \omega \cdot \theta
\end{equation}
where $\omega$ is the cycle mapping factor, set to 2 in this work.

Let $g(\gamma)$ denote the object periodic mapping function. According to the Fourier theorem, any periodic function can be decomposed into an infinite sum of orthogonal trigonometric components. In our FSC framework, we encode the angle by predicting a truncated set of these components. The network outputs a $(2N+1)$-dimensional vector, where $N$ represents the maximum frequency order. The theoretical Fourier series expansion is formulated as:
\begin{equation} \label{encode}
	g(\gamma) \approx f(\gamma) = \frac{a_0}{2} + \sum_{n=1}^{N} \left[ a_n \cos(n\gamma) + b_n \sin(n\gamma) \right]
\end{equation}  
where $n\omega$ represents the oscillation frequency of the $n$-th harmonic. The term $a_0$ is the Direct Current (DC) component, mathematically representing the average value of the object signal over one fundamental period $T$:
\begin{equation}
	a_0 = \frac{1}{\pi} \int_{-\pi}^{\pi} g(\gamma) d\gamma
\end{equation}

While high-frequency components naturally handle the periodic wrapping, explicitly encoding the DC component $a_0$ serves as an essential global anchor. It provides the neural network with a stable baseline offset, preventing the regression from oscillating wildly around zero and thereby improving the overall convergence stability.

\textbf{Decoding.} To enforce structural stability and focus solely on the angular phase, the amplitude coefficients $a_n$ and $b_n$ in Eq. (\ref{encode}) are both set to 1 by default. Thus, after polar coordinate mapping, the single-frequency FSC $f({\hat \theta }) $ is as follows:
\begin{equation}
	z = f({\hat \theta }) = e^{jk\omega{\hat \theta }} = \cos(k\omega{\hat \theta }) + j \sin(k\omega{\hat \theta })
\end{equation}
where $k$ denotes the harmonic order of the Fourier series. Unlike heuristic decoding processes, the orthogonal nature of these Fourier components allows the angle to be exactly and continuously recovered using the standard inverse tangent function. Ultimately, FSC's decoding result $\hat \theta$ can be formulated as:
\begin{equation} \label{decode}
	\begin{aligned}
		{\hat \theta } &= f^{-1}(z) = -\frac{j}{k\omega} \ln z = \frac{1}{k\omega}(\mathrm{arctan2}(\operatorname{Im}(z), \operatorname{Re}(z))) 
	\end{aligned}
\end{equation}
where $\operatorname{Im}(z)$ and $\operatorname{Re}(z)$ represent the imaginary and real parts of a complex number, respectively. $\operatorname{arctan2}$ is a variant of the arctangent function, typically ranging $(-\pi, \pi]$.
\begin{figure}[t]
	\centering{\includegraphics[width=\columnwidth]{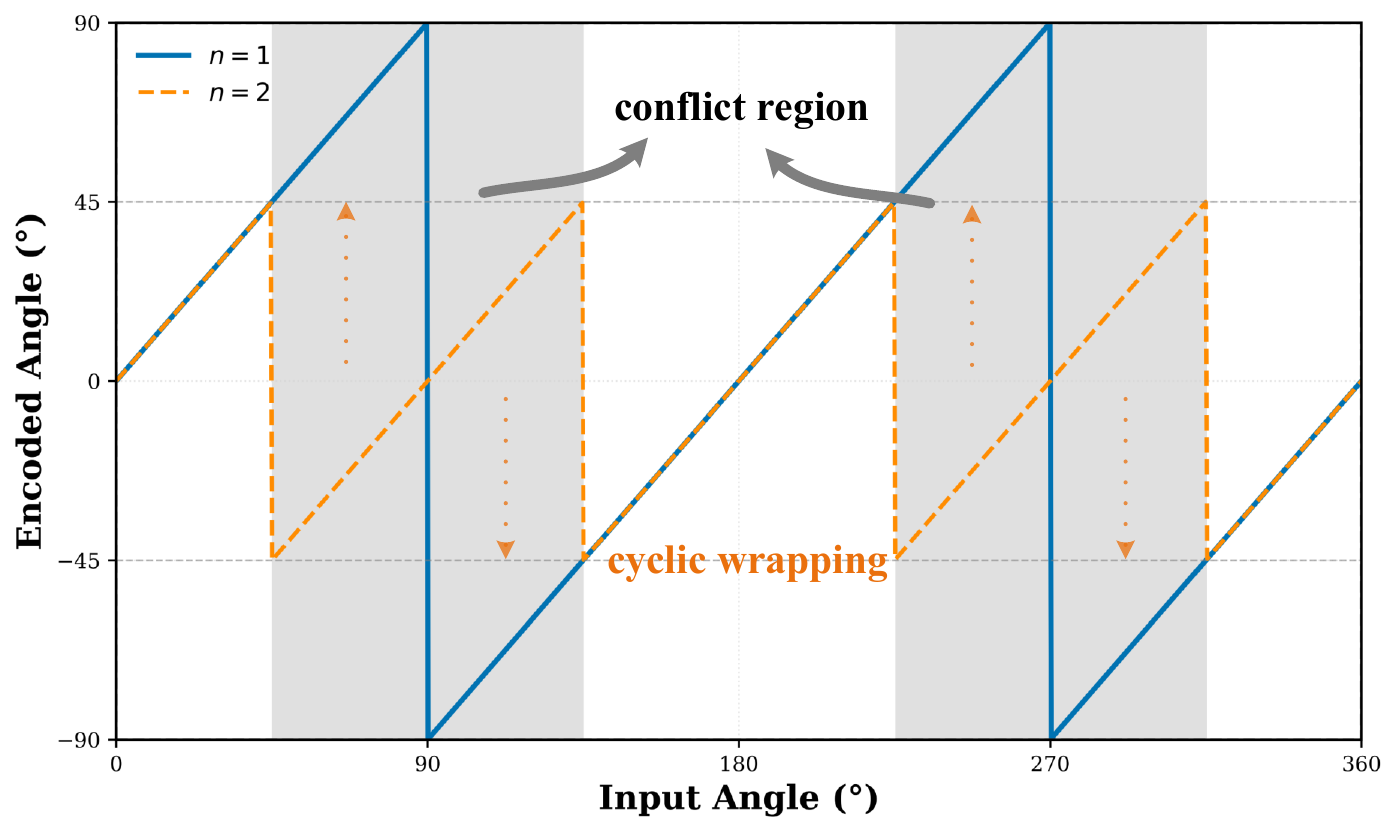}}
	\caption{The angle mapping functions for rectangular and square objects. Our FSC utilizes multi-frequency components to eliminate ambiguity by cyclic wrapping. Conflict regions are marked in gray.}
	\label{cycle_wrap}
\end{figure}

The decoding process is the inverse function of the encoding process. To simplify, Eq. (\ref{decode}) is used for decoding.
Assuming the FSC encoding uses at least two distinct frequency components (e.g., $k=1$ and $k=2$), we have $z_{1} = e^{j\omega{\hat \theta }}$ and $z_{2} = e^{2j\omega{\hat \theta }}$. 
The respective angle estimates corresponding to these two complex values can be formulated as follows:
\begin{equation}
	\begin{gathered}
		\hat{\theta}_{1} = \frac{1}{\omega} {arctan2}(\operatorname{Im}(z_1), \operatorname{Re}(z_1)),	\\
		\hat{\theta}_{2} = \frac{1}{2\omega} {arctan2}(\operatorname{Im}(z_2), \operatorname{Re}(z_2)) 	
	\end{gathered}
\end{equation}

Mapping the aforementioned complex space to polar coordinate space, the decoding process yields vector components $z(x)$ and $z(y)$ on the x-axis and y-axis.  
Consequently, the decoding cycles of $\hat{\theta}_{1}$ and $\hat{\theta}_{2}$ differ by two times, corresponding to the rotation cycles of rectangular and square bounding boxes, respectively. In FSC, $\hat{\theta}_{1}$ and $\hat{\theta}_{2}$ are independent angular deviations only constrained by the Fourier series. Therefore, cyclic wrapping is employed to rectify cyclic ambiguity.
As shown in Fig. \ref{cycle_wrap}, the range of $\hat{\theta}_{2}$ is $( - \pi /4,\pi /4]$, and there is a cyclic deviation when predicting a rectangular bounding box through $\hat{\theta}_{2}$. 
The final prediction $\theta _{{{pred}}}$ can then be formulated as follows:
\begin{equation}
	{\theta _{{{pred}}}} = \begin{cases}
		\frac{({{\hat \theta }_2}\bmod 2\pi ) - \pi }{2} {\rm{ }} &, \text{if } cos(\hat{\theta}_1 - \hat{\theta}_2) < 0 \\
		\frac{\hat{\theta}_2}{2}&, \text{otherwise}
	\end{cases}
	\label{cos}
\end{equation}

Fundamentally, this multi-frequency decoding explicitly models the geometric properties of different objects. Specifically, rectangular objects exhibit $\pi$ periodicity, requiring only the fundamental frequency (n=1), whereas square-like objects exhibit $\pi/2$ periodicity, necessitating the dual-frequency (n=2) to completely resolve ambiguity.
\subsection{Noise Immunity}
\label{sec:Immunity} 

\textbf{Core Definitions and Premises.} To theoretically evaluate the decoding robustness, we model the inherent predictive uncertainty of the neural network as independent Gaussian noise injected into the encoded components. By injecting Gaussian noise into the decoding stage, we provide a detailed analysis of FSC's superior noise immunity compared to state-of-the-art angle coders (e.g., PSC \cite{yu2023phase}).
FSC (n=2) encodes sine and cosine components of the single frequency ($\gamma$) and double frequency ($2\gamma$), resulting in 4 effective channels. It is worth noting that the DC component $a_0$ of the FSC only serves as a supervisory signal during the training stage. 
Thus, adding noise $|\varepsilon|\ll 1$ to the encoding component, the noisy components can be formulated as:
\begin{equation}
	\begin{gathered}
		\cos\gamma_{\text{noisy}} = \cos2\theta + \varepsilon_{f1}, \quad \sin\gamma_{\text{noisy}} = \sin2\theta + \varepsilon_{f2}, \\
		\cos2\gamma_{\text{noisy}} = \cos4\theta + \varepsilon_{f3}, \quad \sin2\gamma_{\text{noisy}} = \sin4\theta + \varepsilon_{f4}
	\end{gathered}
\end{equation}
where $\varepsilon_{f1},\varepsilon_{f2},\varepsilon_{f3},\varepsilon_{f4} \sim \mathcal{N}(0, \sigma_F^2)$ and are independent and identically distributed (i.i.d.).

PSC \cite{yu2023phase} encodes 3-phase cosine components of the single frequency ($\gamma$) and double frequency ($2\gamma$), resulting in 6 effective channels. The cosine noisy components can be formulated as:
\begin{equation}
	\begin{gathered}
		P_{k,\text{noisy}} = \cos\left(2\theta + \frac{2k\pi}{3}\right) + \varepsilon_{p_{k1}} \quad (k=0,1,2), \\
		P_{k2,\text{noisy}} = \cos\left(4\theta + \frac{4k\pi}{3}\right) + \varepsilon_{p_{k2}} \quad (k=0,1,2)
	\end{gathered}
\end{equation}
where $\varepsilon_{p_{k1}}, \varepsilon_{p_{k2}} \sim \mathcal{N}(0, \sigma_P^2)$ are i.i.d.

\textbf{Fluctuation Performance.} The decoding angular deviation, $\mathbb{E}[\Delta\theta]$, is used as the noise immunity metric, where $\Delta\theta = |\hat{\theta} - \theta|$ ($\hat{\theta}$ is the decoding result, $\theta$ is the true angle). $\mathbb{E}[\cdot]$ represents the expected value of the noise distribution. A smaller expected value indicates stronger noise immunity.
The cyclic wrapping solves the cyclic ambiguity problem among multiple frequencies in theory, but it overlooks the decision boundary's sensitivity in Eq. (\ref{cos}). The aforementioned issue involves false-positive and false-negative judgments caused by noise from components. To decode with multi-frequency, we first apply the double-frequency constraint to correct noise deviation. For the single frequency ($\gamma$) and its double frequency ($2\gamma$), we derive their noisy deviation as follows: 
\begin{equation}
	\begin{split}
		\hat{\phi}_{f1} &= \arctan2\left(\sin\gamma_{\text{noisy}}, \cos\gamma_{\text{noisy}}\right), \\
		\hat{\phi}_{f2} &= \frac{1}{2}\arctan2\left(\sin2\gamma_{\text{noisy}}, \cos2\gamma_{\text{noisy}}\right)
	\end{split}
\end{equation}

Assume that the simulated model introduces a minimal prediction deviation, manifested as cosine/sine noise terms where $|\varepsilon_c|, |\varepsilon_s| \ll 1$. 
Thus, $\arctan2(s + \varepsilon_s, c + \varepsilon_c)$ can be obtained by Taylor expansion, which is:
\begin{equation}
	\arctan2(s+\varepsilon_s, c+\varepsilon_c) \approx \arctan2(s,c) + \varepsilon_s \cdot c - \varepsilon_c \cdot s
\end{equation}
where the partial derivatives are $\frac{\partial \text{arctan2}(s,c)}{\partial s} = c$ and $\frac{\partial \text{arctan2}(s,c)}{\partial c} = -s$, respectively, given the orthogonal condition $c^2 + s^2 = 1$.
Consequently, the resulting component deviation $\Delta\phi_{f2}$ is approximated using the differential form:
\begin{equation}
	\Delta\phi_{f2}\approx \frac{1}{2}(\varepsilon_{f_{4}} \cos4\theta - \varepsilon_{f_{4}} \sin4\theta)
\end{equation}

Given the predictive noise ($\text{Var}(\varepsilon_f) = \sigma_F^2$), the variance of the decoded angle deviation $\Delta\phi_{f2}$ evaluates directly to:
\begin{equation}
	\text{Var}(\Delta\phi_{f2}) = \frac{\sigma_F^2}{4} (\cos^2 4\theta + \sin^2 4\theta) = \frac{1}{4} \sigma_F^2.
\end{equation}

Therefore, the fluctuation variance of FSC, derived from $\Delta\theta_{\text{FSC}} = \Delta\phi_{f2}/2$, is:
\begin{equation}
	\text{Var}(\Delta\theta_{\text{FSC}}) = \text{Var}\left(\frac{\Delta\phi_{f2}}{2}\right) = \frac{1}{4} \text{Var}(\Delta\phi_{f2}) = \frac{\sigma_F^2}{16}
\end{equation}

PSC \cite{yu2023phase} adopts a similar strategy for dual-frequency decoding. However, since it constructs the Cartesian coordinates from a non-orthogonal 3-phase basis ($\alpha_k = 0, \frac{2\pi}{3}, \frac{4\pi}{3}$), its synthesized components $S = -\sum P_{k2} \sin\alpha_k$ and $C = \sum P_{k2} \cos\alpha_k$ inherently amplify the signal, resulting in a squared modulus of $C^2 + S^2 = \frac{9}{4}$. Following a similar Taylor expansion for this non-orthogonal projection, the predictive noise propagates through the scaled amplitude. Consequently, the variance of the double-frequency deviation $\Delta\phi_{p2}$ evaluates to:

\begin{equation}
	\text{Var}(\Delta\phi_{p2}) = \frac{1}{4} \left( \frac{4}{9} \cdot \frac{3}{2} \sigma_P^2 \right) = \frac{1}{6} \sigma_P^2.
\end{equation}

Therefore, the fluctuation variance of PSC, derived from $\Delta\theta_{\text{PSC}} = \Delta\phi_{p2}/2$, is determined as:
\begin{equation}
	\text{Var}(\Delta\theta_{\text{PSC}}) = \left( \frac{1}{2} \right)^2 \cdot \frac{1}{6}\sigma_P^2 = \frac{\sigma_P^2}{24}
\end{equation}

\textbf{Discussion on Noise.} To evaluate the decoding stability, we first analyze the hypothetical condition where the initial predictive noises of both coders' components are equivalent ($\sigma_F = \sigma_P = \sigma$). Under this ideal mathematical premise, the variance of the decoding angle deviation for FSC is $\frac{1}{16}\sigma^2$, while PSC achieves a smaller variance of $\frac{1}{24}\sigma^2$. This indicates that theoretically, the 3-phase overcomplete design of PSC possesses a slight advantage in noise tolerance if the base neural noises are identical.
However, this ideal assumption is difficult to strictly maintain in practical neural network regression. As demonstrated in the Monte Carlo simulation in Fig. \ref{pic2}\textcolor{red}{(c)}, PSC's non-orthogonal projection mechanism lacks explicit manifold constraints. Without such structural regularization, the network may experience modulus degradation during optimization, where the synthesized Cartesian amplitude tends to decay. When the modulus decreases, the relative impact of the actual regression noise $\sigma_P$ on the signal is amplified. This relative noise inflation can offset PSC's theoretical advantage, potentially leading to cycle errors observed in Fig. \ref{pic2}\textcolor{red}{(d)}.
In contrast, FSC employs a constrained, minimal orthogonal Fourier basis. By regularizing the predictions toward the unit circle, it helps maintain a stable $\sigma_F$. Consequently, FSC effectively mitigates the risk of noise inflation, translating its robust mathematical manifold into improved practical noise immunity.
\subsection{Loss Function}
\label{sec:Loss}

During the training stage, the FSC performs additional encoding on the angle. In the testing stage, the angle decoded by the FSC is used as the prediction result.
An additional Fourier series regression feature $F_{fourier}={{\mathbb{R}}^{H\times W\times (2N+1)}}$ is output from the regression feature $F_{reg}={{\mathbb{R}}^{H\times W\times 256}}$, which can be formulated as follows:
\begin{equation}
	{{F}_{{fourier}}} = 2 \times {sig}moid({F_{reg}}) - 1
\end{equation}
where $2N+1$ is the encoding length of FSC. $sigmoid$ is a normalization function that constrains feature values to the range of $[0,1]$. Therefore, the feature values of ${F}_{fourier}$ conform to the value range $[- 1,1]$ of the FSC encoding components. As illustrated in Fig. \ref{main}, we calculate the loss of the angular encoding in FSC, which is
\begin{equation}
	\begin{split}
		L_{fsc} = & \underbrace{ \sum\limits_{i = 1}^{2N+1} L_{reg}(\tilde{\theta}_i^{pred}, \tilde{\theta}_i^{gt}) }_{\text{Target Fitting}} \\
		& + \underbrace{ \sum\limits_{k = 1}^N L_{reg}\Big( (\tilde{\theta}_{2k}^{pred})^2 + (\tilde{\theta}_{2k+1}^{pred})^2, 1 \Big) }_{\text{Manifold Constraint}}
	\end{split}
\end{equation}
where the loss is normalized by the number of positive samples $N_{pos}$. $\tilde{\theta}_i^{pred}$ and $\tilde{\theta}_i^{gt}$ represent FSC's predicted and ground-truth encoding components, respectively. $\tilde{\theta}_{2k}$ and $\tilde{\theta}_{2k+1}$ naturally represent the predicted cosine and sine components for the $k$-th frequency. When $L_{reg}$ uses the smooth L1 Loss, the final total loss can be formulated as:
\begin{equation}
	{L_{total}} = {\lambda _1}{L_{box}} + {\lambda _2}{L_{fsc}} + {L_{cls}}
\end{equation}
where $L_{cls}$ is the classification loss, which employs Focal Loss \cite{lin2017focal}. $\lambda_1$ and $\lambda_2$ are set to 1.0 and 0.2 by default, respectively.

\section{Experiment}

\subsection{Datasets}
\label{Experiment}
\textbf{HRSC-2016} \cite{liu2017high} comprises 1,061 high-resolution images collected from six major ports and annotated with oriented bounding boxes. It is divided into training (436 images), validation (181 images), and test (444 images) sets.

\textbf{DOTA-v1.0} \cite{xia2018dota} comprises 2,806 aerial images with varying resolutions ($800\times800$ to $4000\times4000$ pixels) collected from diverse sensors and platforms. It contains 188,282 instances across 15 categories: Plane (PL), Baseball diamond (BD), Bridge (BR), Ground track field (GTF), Small vehicle (SV), Large vehicle (LV), Ship (SH), Tennis court (TC), Basketball court (BC), Storage tank (ST), Soccer ball field (SBF), Roundabout (RA), Harbor (HA), Swimming pool (SP), and Helicopter (HC). In our experiments, we divide images into 1024x1024 sub-images with a 200-pixel overlap and apply random horizontal, vertical, and diagonal flips during training. All our relevant experimental results are from the official evaluation website\footnote{DOTA: \url{http://bed4rs.net:8001/evaluation1/}}.

\textbf{DIOR-R} \cite{cheng2022anchor}, based on DIOR \cite{li2020object} dataset, consists of 23,463 images and 192,518 instances annotated with oriented bounding boxes. It covers 20 common categories: Airplane (APL), Airport (APO), Baseball Field (BF), Basketball Court (BC), Bridge (BR), Chimney (CH), Expressway Service Area (ESA), Expressway Toll Station (ETS), Dam (DAM), Golf Field (GF), Ground Track Field (GTF), Harbor (HA), Overpass (OP), Ship (SH), Stadium (STA), Storage Tank (STO), Tennis Court (TC), Train Station (TS), Vehicle (VE), and Windmill (WM).

\subsection{Experimental Details.} 
\textbf{Training Details.} All experiments are trained using a single NVIDIA 3090 GPU with a batch size of 2 on the DIOR-R, DOTA-v1.0, and HRSC-2016 datasets for 36, 36, and 72 epochs, respectively. 
For the 36-epoch training, the initial learning rate of $2.5 \times {10}^{-2}$ is reduced by a factor of 10 at epochs 24 and 33. For the 72-epoch training, this same reduction is applied at epochs 48 and 66. SGD optimization with weight decay of $1.0 \times  {10}^{-2}$, momentum of 0.9, and gamma of 0.1 is used. 
FSC can seamlessly integrate with any state-of-the-art models, such as ReDet \cite{han2021redet} and ${\text{S}^\text{2}}\text{A-Net}$ \cite{han2021align}. ResNet-50 \cite{he2016deep} is employed as the backbone for all training unless stated otherwise.

\textbf{Evaluation Details.} To ensure fairness and simplicity, all experiments are conducted on the same benchmark, MMRotate \cite{zhou2022mmrotate} without any trick. 
Notably, AP$_{75}$ is more sensitive to deviations in box position and angle, thereby enabling more effective evaluation of performance advantages.
AP$_{75}$ is adopted as main metric, while the others are used as auxiliary metrics.

\begin{table}[h]
	\centering
	\caption{Ablation study with various encoding lengths on FSC.}
	\label{tab:ablation_angle_loss}
	\renewcommand{\arraystretch}{1.2}
	\resizebox{\linewidth}{!}{
		\begin{tabular}{c|c|cc|cc|cc|cc}
			\hline
			\multirow{2}{*}{$a_0$}& \multirow{2}{*}{$cw$} & \multicolumn{2}{c|}{Direct} & \multicolumn{2}{c|}{FSC (n=1)} & \multicolumn{2}{c|}{FSC (n=2)} & \multicolumn{2}{c}{FSC (n=4)} \\
			\cline{3-10}
			&  & $\text{AP}_{50}$ & $\text{AP}_{75}$ & $\text{AP}_{50}$ & $\text{AP}_{75}$ & $\text{AP}_{50}$ & $\text{AP}_{75}$ & $\text{AP}_{50}$ & $\text{AP}_{75}$ \\
			\hline
			-& - & 83.90 & 51.60 & 85.20 & 58.10 & 85.10 & \textbf{48.60} & 83.10& 27.30 \\  
			\checkmark & - & - & - & 86.40 & 58.50 & 85.20 & 47.40 & 82.20 & 26.40 \\  
			\rowcolor{gray!25}
			\checkmark & \checkmark & - & - & \textbf{86.40} & \textbf{58.50} & \textbf{85.60} & 48.10 & \textbf{84.30} & \textbf{28.80} \\  
			\hline
		\end{tabular}
	}
\end{table}

\subsection{Ablation Studies} 

	\textbf{Different Encoding Length $n$.} To demonstrate the effectiveness of encoding length $n$ and the DC component $a_0$, we performed ablation experiments on the HRSC-2016 dataset, as shown in Tab. \ref{tab:ablation_angle_loss}. 
	We find that implementing $a_0$ in loss supervision consistently improves the overall $\text{AP}_{50}$.
	The cyclic wrapping $cw$ mechanism applies dual-frequency for cyclic correction, defaulting to the single-frequency for decoding otherwise.
	The results in the last two rows of Tab. \ref{tab:ablation_angle_loss} indicate that $cw$ effectively offsets partial prediction deviation, and the comprehensive performance comparison between $n=1$ and $n=2$ conclusively demonstrates that FSC effectively addresses boundary discontinuities.
	
	Regarding oriented object detection, objects typically exhibit either rectangular or square shape distributions. 
	For the HRSC-2016 dataset with only rectangular objects, FSC (n=1) achieved the best $\text{AP}_{75}$ and $\text{AP}_{50}$ of 86.40\% and 58.50\% respectively.
However, we observe a theoretical degradation in performance when extending to higher-order frequencies ($n \ge 3$). As indicated by our Taylor expansion analysis in Sec. \ref{sec:Immunity}, the rapid oscillation of high-frequency components inherently amplifies base neural predictive noise during the inverse tangent decoding process. This imposes stringent constraints on the network's regression capacity, suggesting that a low-frequency orthogonal basis ($n=1, 2$) provides the optimal structural balance between resolving ambiguity and maintaining noise immunity.

	\begin{figure}[t]
		\centering
		\subfloat[Angular Error Distribution. \label{error_scatter_3way}]{%
			\includegraphics[width=0.5\linewidth]{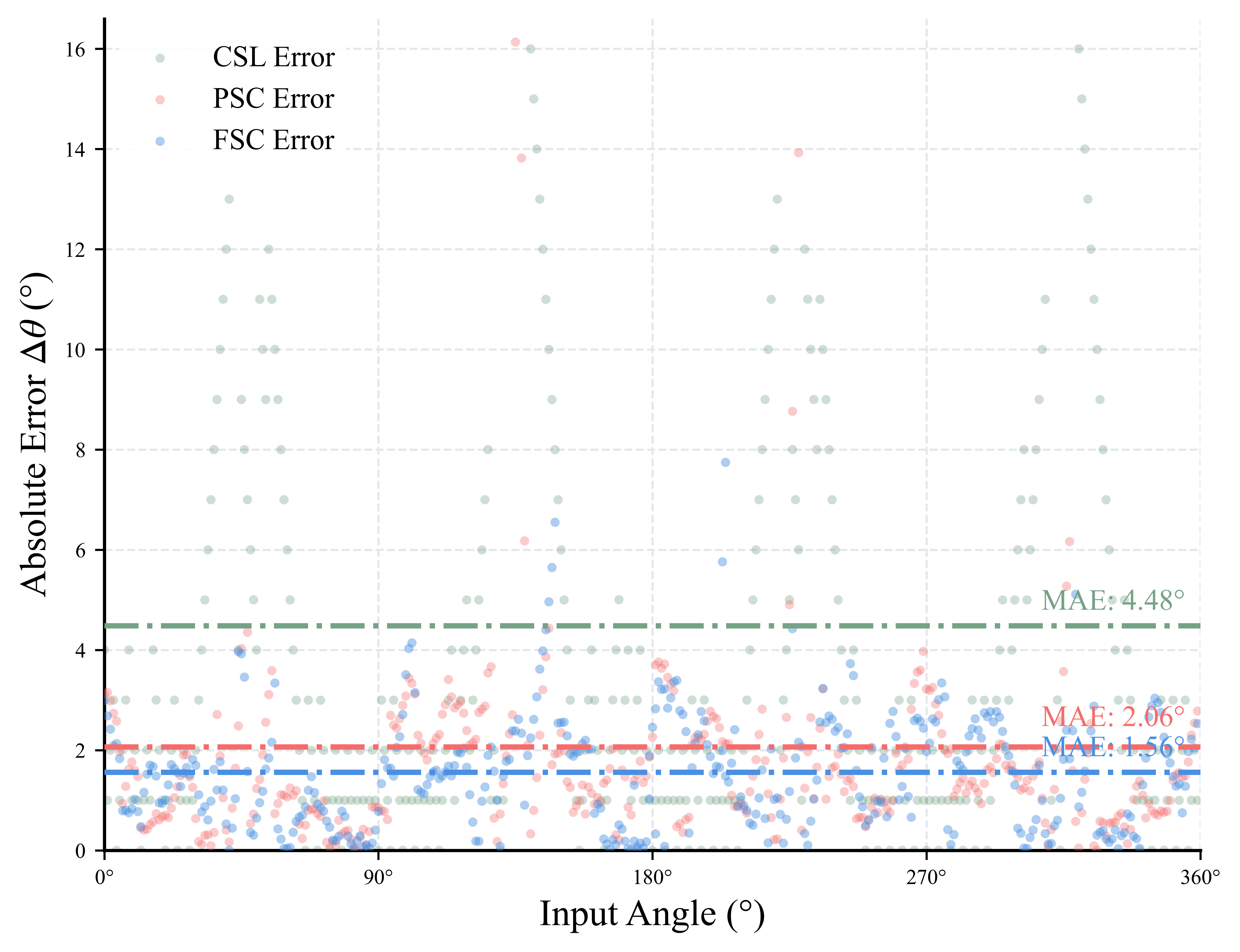}%
		}
		\hfil
		\subfloat[Cumulative Error Distribution. \label{error_cdf_3way}]{%
			\includegraphics[width=0.5\linewidth]{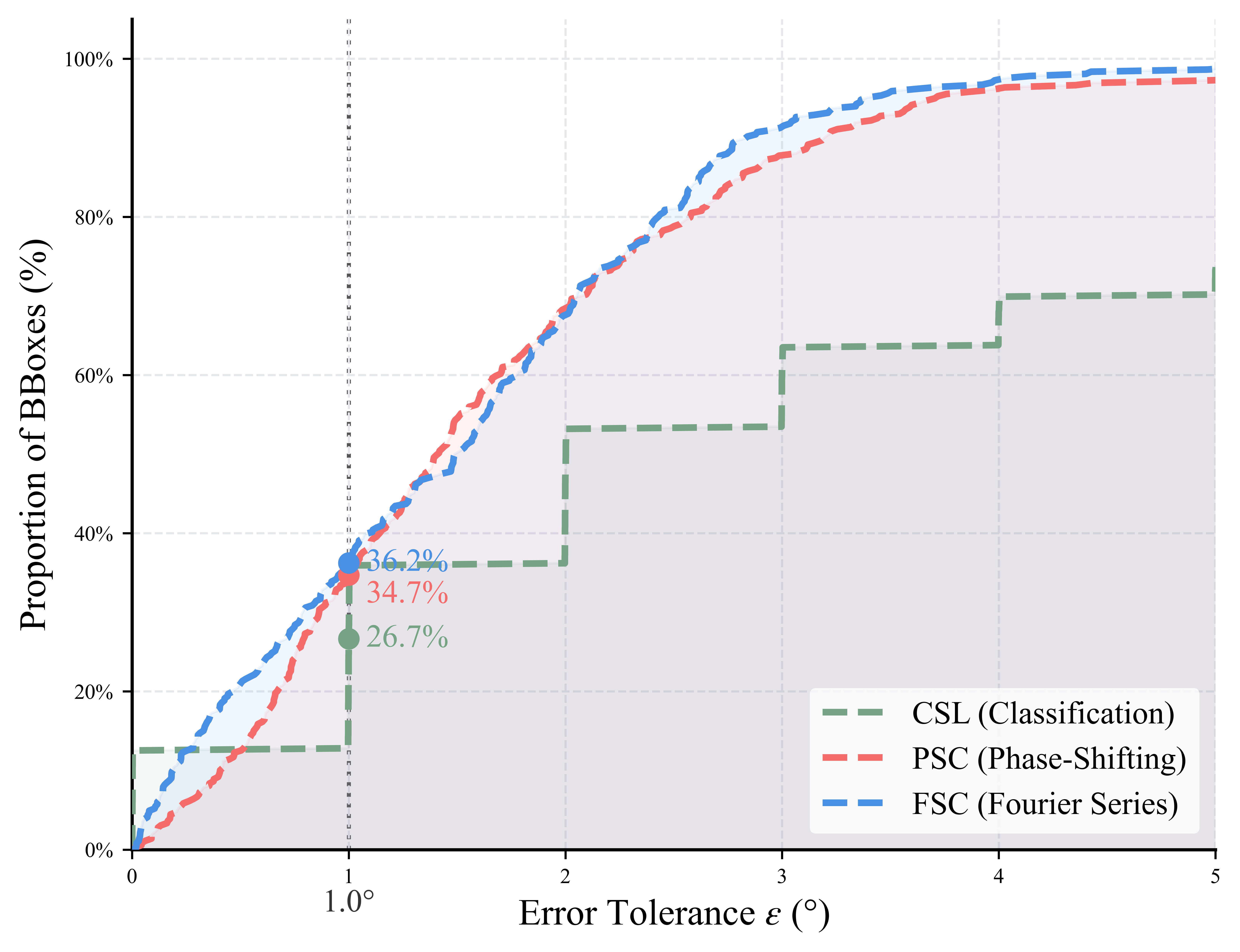}%
		}
		\caption{Compare the angle prediction errors of three angle encoders: FSC (n=1), PSC, and CSL.}
		\label{error_3way}
	\end{figure}
\textbf{Manifold Constraint.} To rigorously validate the effectiveness of the proposed manifold constraint, we conduct both fine-grained angular error analyses and comprehensive metric evaluations. We first compare the angle prediction errors among three coders (CSL, PSC, and FSC) under a single-frequency setting. As illustrated in Fig. \ref{error_3way}\textcolor{red}{(a)}, the classification-based CSL exhibits a scattered error distribution due to its inherent quantization limitations. Both PSC and FSC effectively mitigate this issue, demonstrating highly compact error distributions.

	To further quantify the sub-degree regression performance, Fig. \ref{error_3way}\textcolor{red}{(b)} visualizes the Cumulative Error Distribution (CDF). Under a strict error tolerance of $\Delta\theta \le 1^\circ$, the proportion of accurate bounding boxes for CSL is merely 26.7\%. In contrast, the continuous regression mechanisms showcase a significant advantage. Benefiting from the minimal orthogonal basis, FSC further outperforms the unconstrained PSC (36.2\% vs. 34.7\%), indicating that the geometric regularization provides sharper gradient guidance for fine-grained alignment.
	
	Beyond the statistical error distribution, we evaluate the direct impact of the manifold constraint on internal decoding stability. Tab. \ref{tab:pscd_ablation} presents a quantitative ablation study on the DOTA-v1.0 dataset, utilizing the RetinaNet baseline equipped with PSCD/FSC (n=2). 
	The unconstrained PSCD baseline achieves a competitive $\text{AP}_{50}$ of $71.09\%$. However, an analysis of the underlying prediction metrics reveals a notable limitation: the unconstrained encoder exhibits a substantial component deviation ($\text{MAE}_c = 0.79$), which leads to a significantly high decoded angle error ($\text{MAE}_d = 56.95^\circ$). This phenomenon suggests that a moderately high AP value does not necessarily guarantee precise angular regression, as this metric can partially mask cycle errors induced by feature modulus collapse.
	
		\begin{table}[t]
		\centering
		\small
		\caption{Performance comparison of PSCD w/ and w/o manifold constraints. $\text{MAE}_c$ and $\text{MAE}_d$ denote the Mean Abso lute Error of the raw predicted components and the final decoded angles, respectively. }
		\label{tab:pscd_ablation}
		\begin{tabular}{lcccc}
			\toprule
			Manifold Constraint& MAE$_c \downarrow$ & MAE$_d \downarrow$ & AP$_{50} \uparrow$ & AP$_{75} \uparrow$ \\
			\midrule
			PSCD (w/o) & 0.79 & 56.95 & 71.09 & 41.17 \\
			PSCD (w/)  & 0.38 & 45.12 & \textbf{72.87} &  \textbf{45.98}     \\
			FSC (n=2) (w/o) & 0.36 &  52.23& 71.47  & 44.99 \\
			FSC (n=2) (w/)  & \textbf{0.35} & \textbf{44.82} & 71.92 &  44.99    \\
			\bottomrule
		\end{tabular}
	\end{table}
	\begin{table}[t]
		\centering
		\caption{Ablation with different weights of angle loss on FSC.}
		\label{tab:ablation_angle_loss1}
		\resizebox{\linewidth}{!}{
			\begin{tabular}{cc|ccc|ccc}
				\hline
			\multirow{2}{*}{$\lambda_1$} & \multirow{2}{*}{$\lambda_2$} & \multicolumn{3}{c|}{FSC (n=1)} & \multicolumn{3}{c}{FSC (n=2)} \\
			\cline{3-8}  
			&  & AP$_{50}$ & AP$_{75}$ & AP & AP$_{50}$ & AP$_{75}$ & AP \\  
			\hline 
			1.0 & 0.0 & 83.90 & 51.60 & 50.41 & - & - & - \\
			0.5 & 0.5 & 86.00 & \textbf{60.20} & \textbf{53.34} & 85.40 & \textbf{49.50} & \textbf{48.83} \\
			0.6 & 0.6 & 86.30 & 50.00 & 49.70 & 85.60  & 45.90  & 47.70 \\
			0.7 & 0.7 & 85.90 & 57.60 & 52.39 &\textbf{85.80} & 41.20 & 46.03 \\
			\rowcolor{gray!25}
			0.7 & 0.6 & \textbf{86.40} & 58.50 & 53.04& 85.60 & 48.10  & 48.01 \\
			0.8 & 0.8 & 85.70 & 58.00 & 52.50 & 85.40 & 48.40 & 48.15 \\
			0.9 & 0.9 & \textbf{86.40} & 50.90  & 50.92& 84.50  & 49.00 & 48.31 \\
			1.0 & 1.0 & 85.50 & 57.40 & 51.93 & 85.20 & 38.40 & 44.26 \\
			\hline
	\end{tabular}}
	\end{table}
	
		\begin{figure}[t]
		\centering
		
		\subfloat[FSC (n=1). \label{fsc1_fresh_scatter_fitting}]{%
			\includegraphics[width=0.5\linewidth]{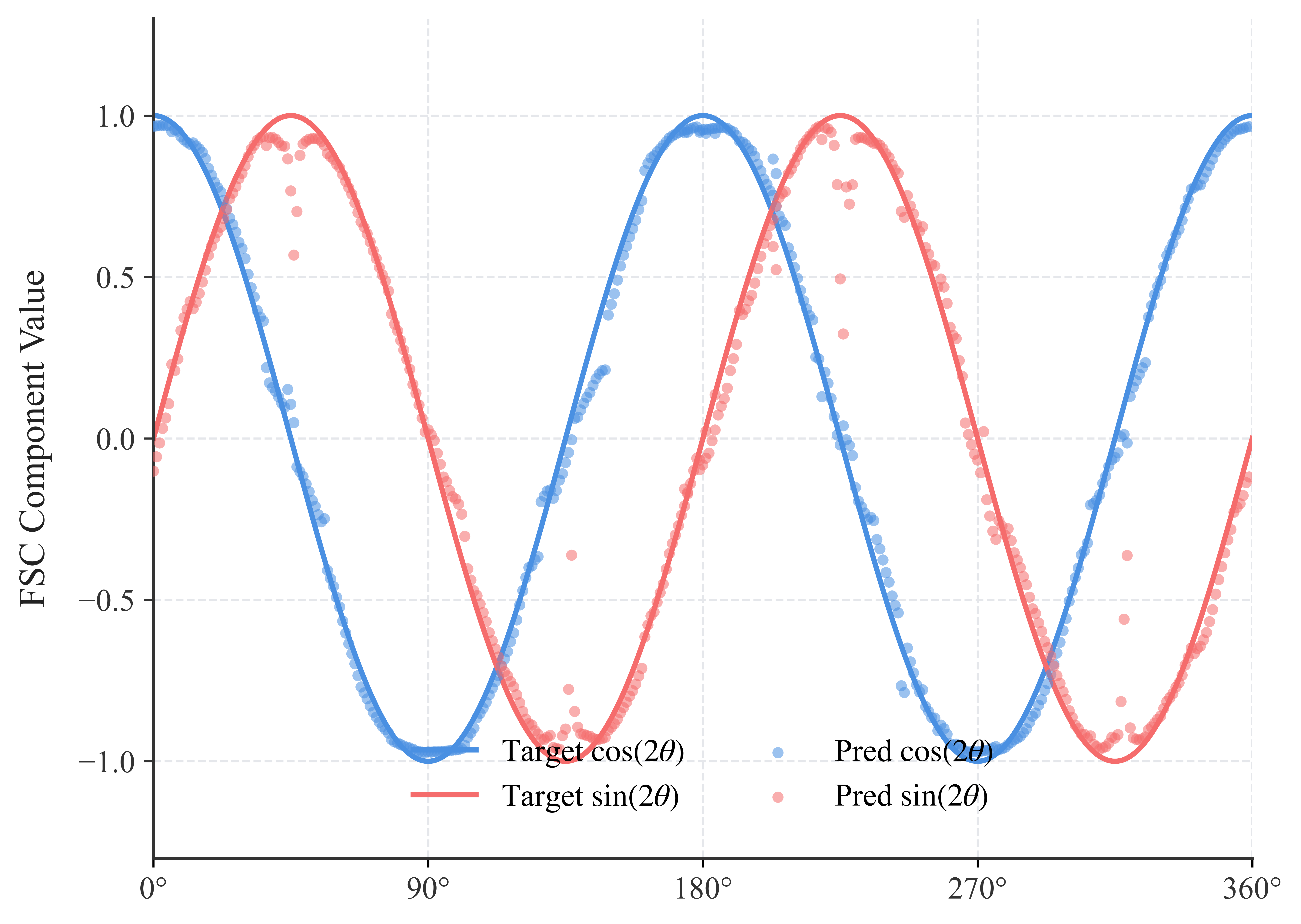}%
		}
		\hfil
		\subfloat[PSC. \label{psc_true_fitting}]{%
			\includegraphics[width=0.5\linewidth]{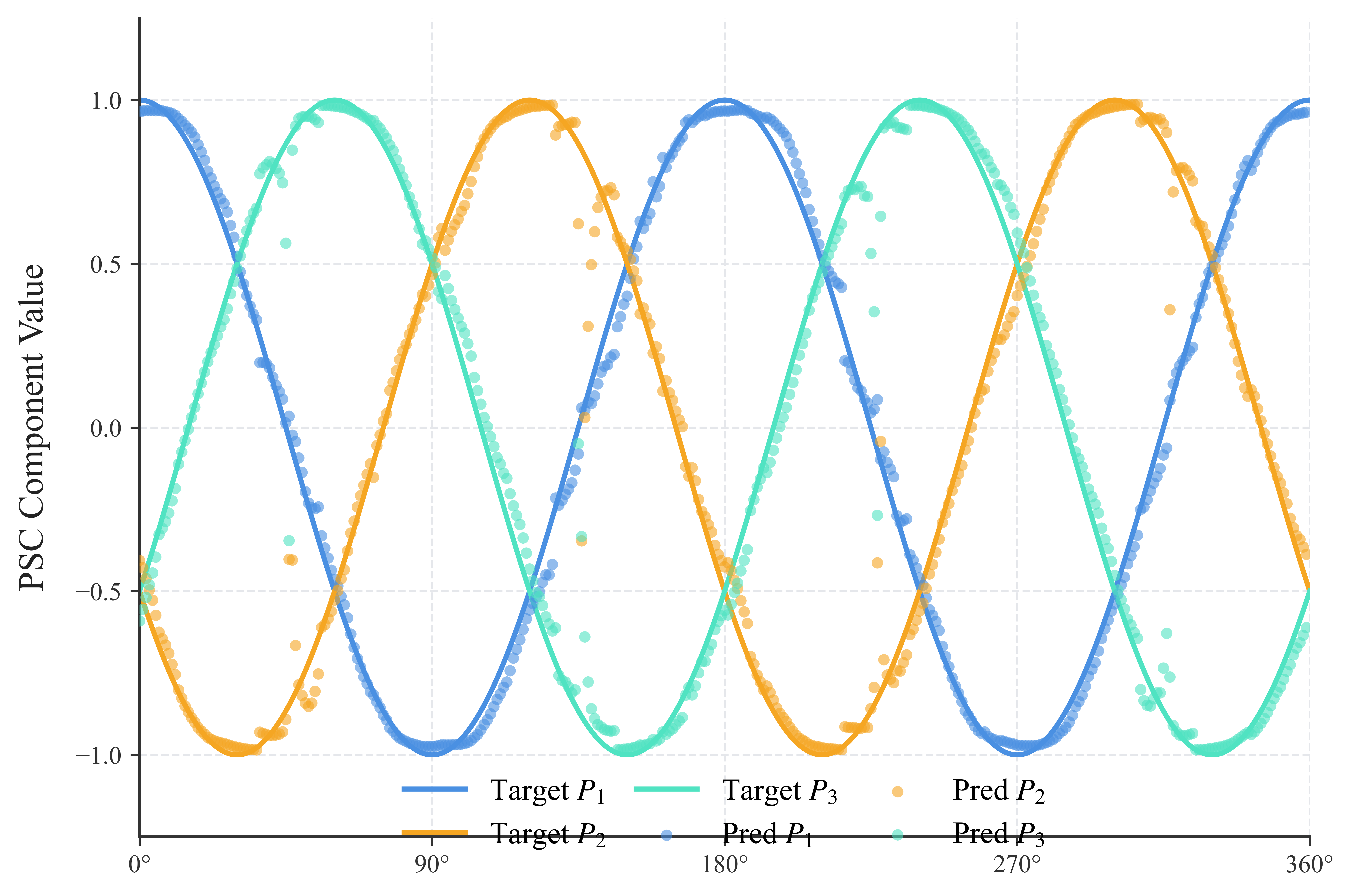}%
		}
		\caption{Comparison of component fluctuations between PSC and FSC (n=1).}
		\label{component fluctuations}
	\end{figure}
	\begin{figure}[t]
		\centering
		\subfloat[Unconstrained.\label{dpsc_components}]{%
			\includegraphics[width=0.5\linewidth]{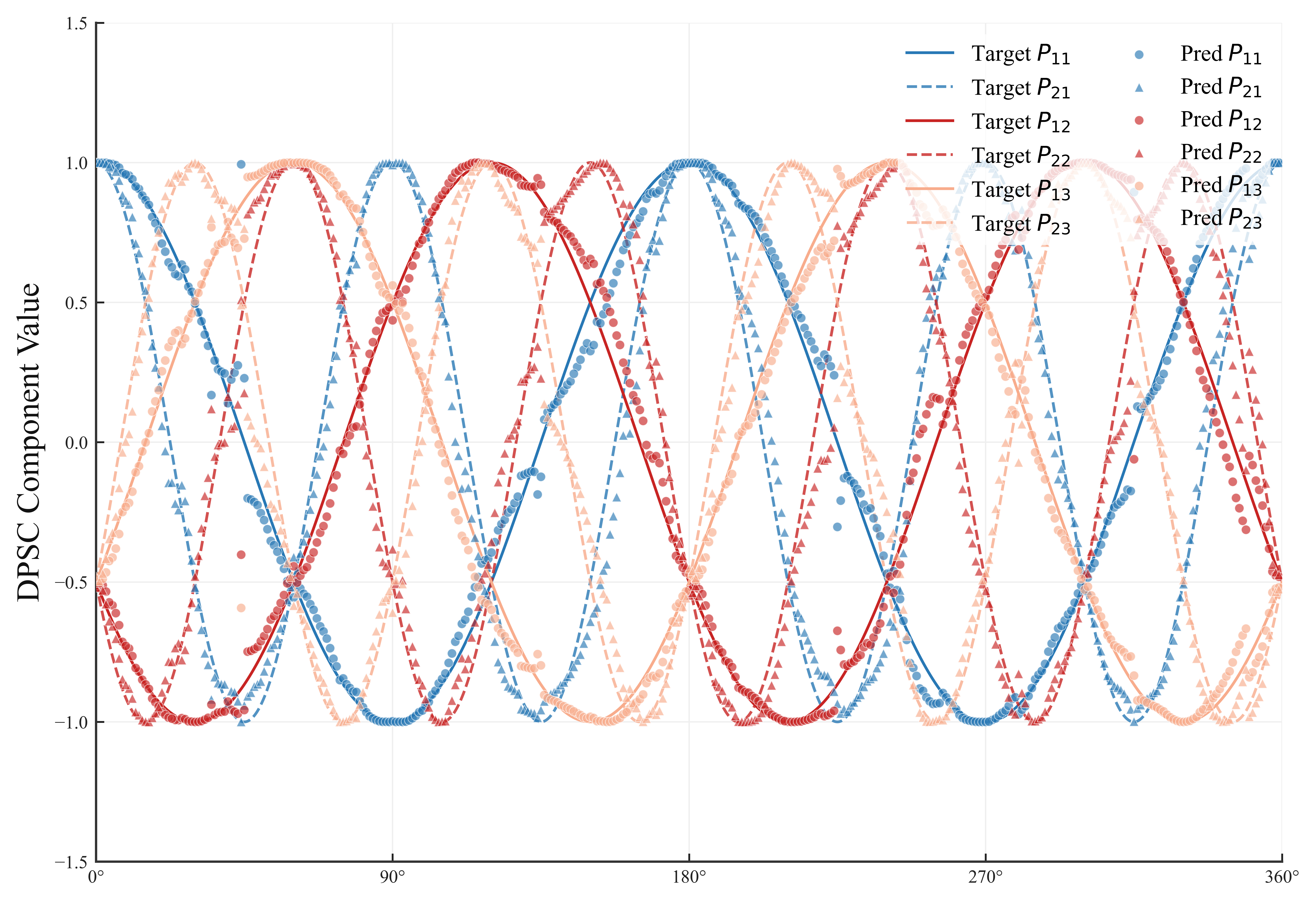}%
		}
		\hfil
		\subfloat[Manifold-Constrained. \label{dpsc_components_loss1}]{%
			\includegraphics[width=0.5\linewidth]{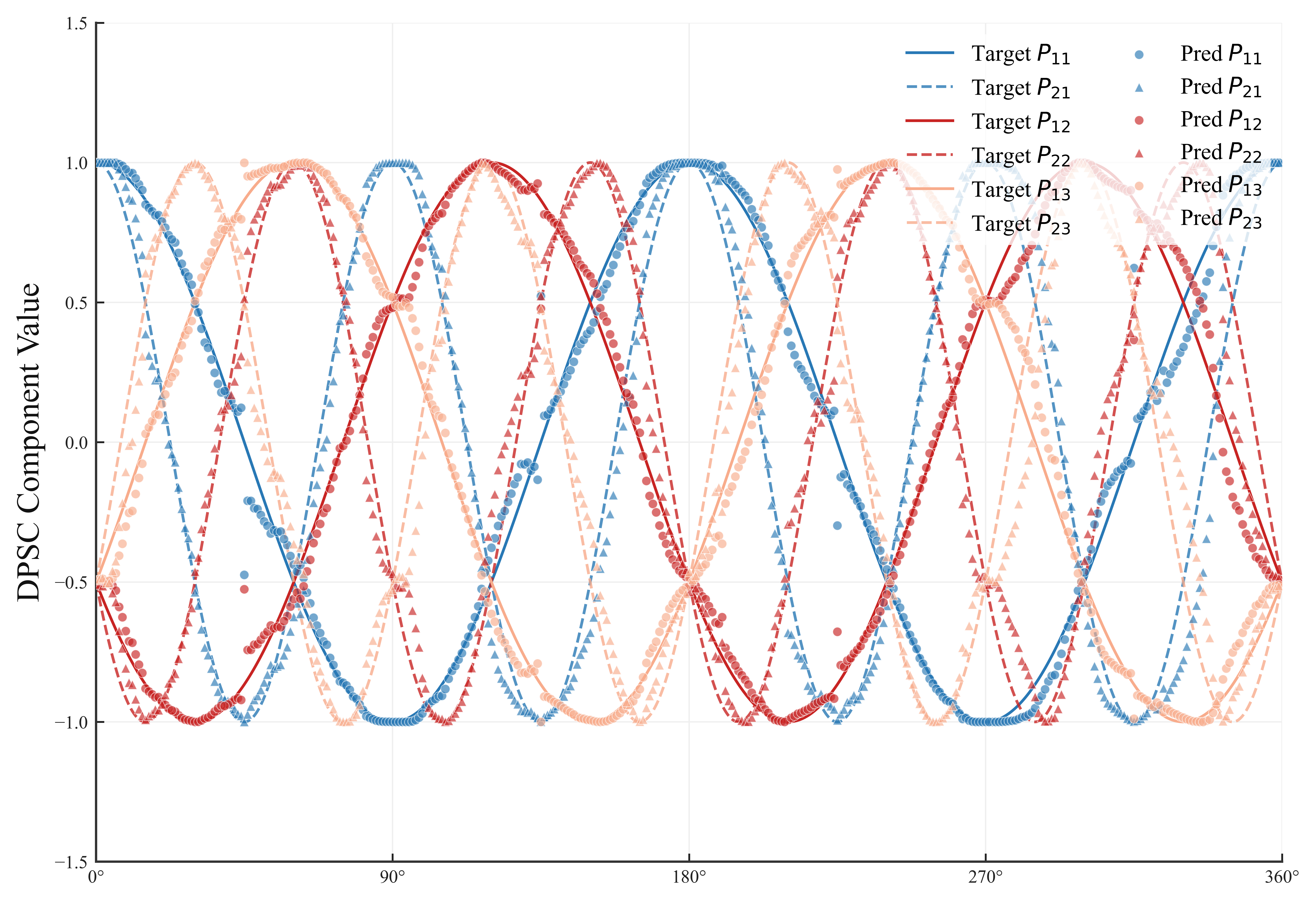}%
		}
		\hfil
		\subfloat[Decoded Angles w/o Constraint. \label{rr_pscd}]{%
			\includegraphics[width=0.5\linewidth]{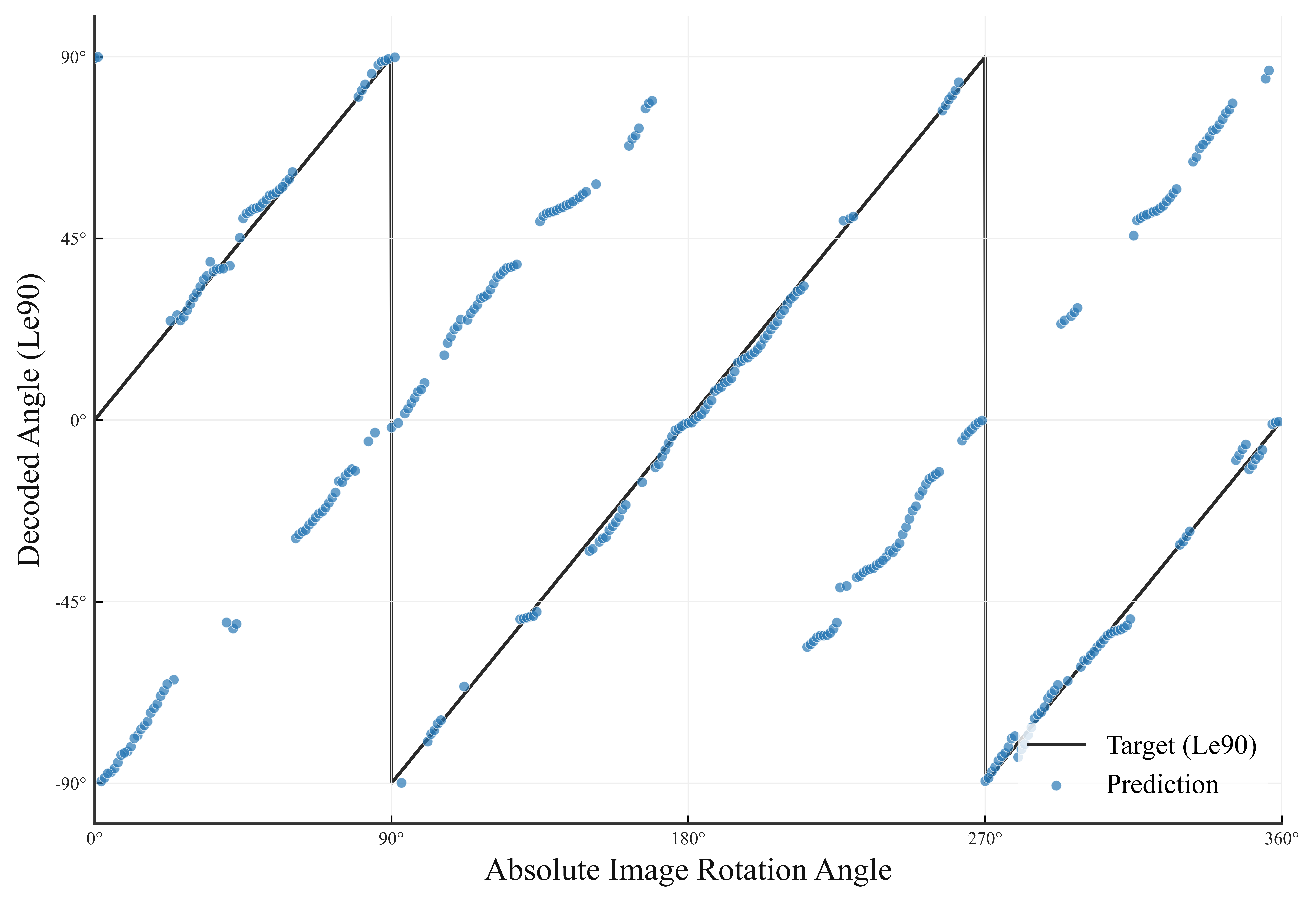}%
		}
		\hfil
		\subfloat[Decoded Angles w/ Constraint. \label{rr_pscd_loss1}]{%
			\includegraphics[width=0.5\linewidth]{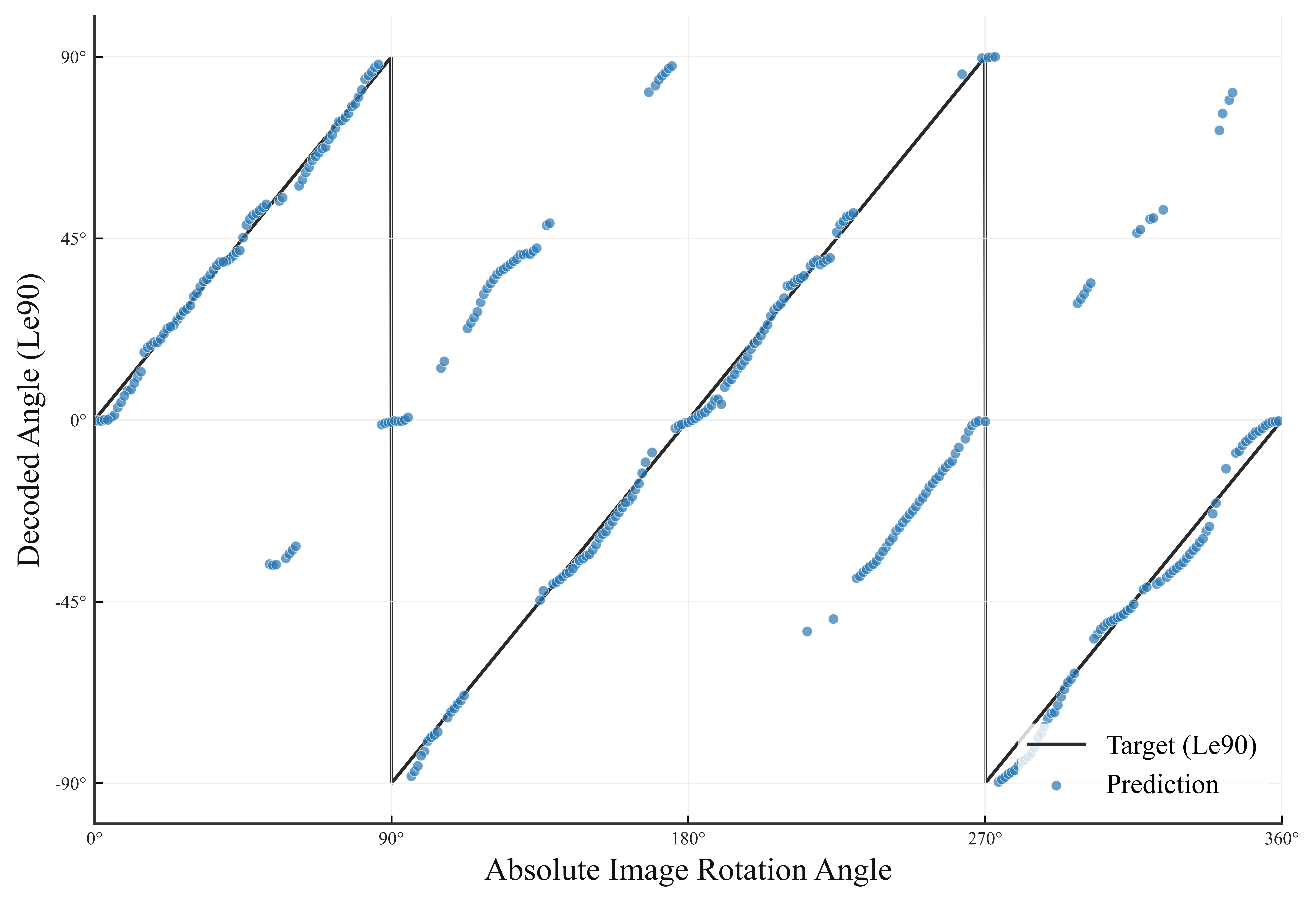}%
		}
		\caption{Visual analysis of the manifold constraint. (a) Raw PSCD components exhibit severe drift. (b) Our constraint locks the features onto the theoretical trigonometric manifold. (c) Unconstrained features lead to fragile decoding and cycle errors for square-like objects. (d) Our constrained representations enable robust decoding.}
		\label{Vis_manifold}
	\end{figure} 
	
	In contrast, the proposed manifold constraint effectively suppresses these internal instabilities. The constrained PSCD reduces $\text{MAE}_c$ to $0.38$ and lowers $\text{MAE}_d$ to $45.12^\circ$. Consequently, this stabilized feature representation contributes to a noticeable performance improvement, elevating the $\text{AP}_{75}$ to $45.98\%$.
	Furthermore, our proposed FSC (n=2) achieves a highly robust angular error ($\text{MAE}_d$=$44.82^\circ$) directly through its minimal orthogonal basis. While the overall $\text{AP}_{75}$ of 44.99 for FSC is marginally lower than that of the constrained PSCD, this empirical outcome perfectly aligns with our theoretical derivations in Sec. \ref{sec:Immunity}. As mathematically proven, when feature modulus collapse is effectively prevented, the three-phase overcomplete design inherently possesses a slightly superior theoretical noise tolerance (variance of $\frac{1}{24}\sigma^2$) compared to the minimal orthogonal Fourier basis (variance of $\frac{1}{16}\sigma^2$). This result comprehensively validates our theoretical framework regarding structural stability and noise immunity.
	
	\begin{figure*}[t]
		\centering{\includegraphics[width=\textwidth]{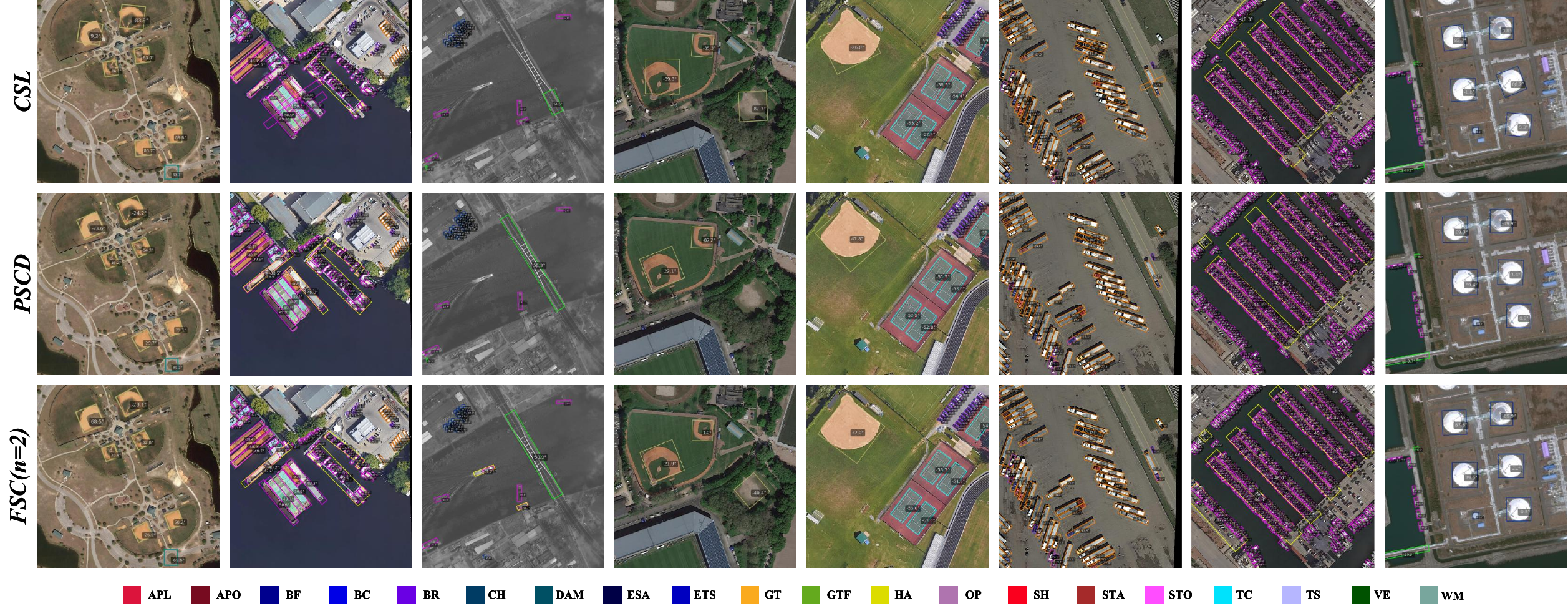}}
		\caption{Visual comparison of three angle coders on the DOTA-v1.0 dataset.}
		\label{vis_dota}
	\end{figure*}
	
	\begin{table*}[t]
		\centering
		\small
		\caption{Comparisons with state-of-the-art OBB/angle coders on \textbf{DOTA-v1.0} dataset. $\dagger$ means using random rotation augmentation.}
		\label{tab:dota_ap50}
		\renewcommand{\arraystretch}{1.2}
		\resizebox{\linewidth}{!}{
			\begin{tabular}{l|c|c|c|c|c|c|c|c|c|c|c|c|c|c|c|c|c|c|c}
				\hline
				Baseline & Angle Coder & PL & BD & BR & GTF & SV & LV & SH & TC & BC & ST & SBF & RA & HA & SP & HC & AP$_{50}$ & AP$_{75}$ & AP \\
				\hline
				\multirow{7}{*}{\rotatebox{90}{\parbox{2cm}{\centering FCOS}}} 
				& Direct & \textbf{89.18} & 71.99 & 47.97 & \textbf{61.61} & 79.30 & 73.52 & 85.78 & 90.90 & 81.09 & 84.30 & \textbf{59.57} & 62.69 & 62.08 & 69.94 & 49.31 & 71.28 & 37.08 & 39.42 \\
				
				& +KLD & 88.96 & 74.42 & 49.41 & 58.22 & 79.85 & 77.75 & 87.02 & \textbf{90.91} & \textbf{82.76} & 84.34 & 56.18 & 64.60 & 63.73 & 68.43 & 45.15 & 71.45 & 39.03 & 40.10 \\

				& +CSL & 88.24 & 74.87 & 41.27 & 61.03 & 79.52 & 78.35 & 87.19 & 90.88 & 81.50 & \textbf{84.53} & 54.70 & 62.65 & 62.84 & 68.45 & 46.50 & 70.83 & 38.71 & 39.75 \\
				& +PSC& 88.69 & 70.34 & 49.82 & 63.14 & 79.52 & 77.96 & 87.08 & 90.89 & 82.53 & 82.88 & 57.39 & 64.17 & 64.10 & 68.85 & 51.77 & 71.94 & 39.74 & 40.79 \\
				
				& +PSCD& 88.49 & 75.53 & 48.10 & 61.50 & 79.93 & 77.13 & 86.92 & 90.90 & 80.68 & 81.44 & 56.15 & \textbf{64.81} & 63.74 & 68.86 & 38.61 & 70.85 & 40.01 & 40.46 \\
				& \multicolumn{1}{>{\columncolor{gray!25}}c}{+FSC (n=1)} &
				\multicolumn{1}{>{\columncolor{gray!25}}c}{88.35} & \multicolumn{1}{>{\columncolor{gray!25}}c}{76.92} & \multicolumn{1}{>{\columncolor{gray!25}}c}{\textbf{50.25}} & \multicolumn{1}{>{\columncolor{gray!25}}c}{61.58} & \multicolumn{1}{>{\columncolor{gray!25}}c}{\textbf{80.04}} & \multicolumn{1}{>{\columncolor{gray!25}}c}{81.01} & \multicolumn{1}{>{\columncolor{gray!25}}c}{87.41} & \multicolumn{1}{>{\columncolor{gray!25}}c}{90.87} & \multicolumn{1}{>{\columncolor{gray!25}}c}{78.85} & \multicolumn{1}{>{\columncolor{gray!25}}c}{76.52} & \multicolumn{1}{>{\columncolor{gray!25}}c}{56.66} & \multicolumn{1}{>{\columncolor{gray!25}}c}{62.92} & \multicolumn{1}{>{\columncolor{gray!25}}c}{\textbf{71.26}} & \multicolumn{1}{>{\columncolor{gray!25}}c}{\textbf{72.68}} & \multicolumn{1}{>{\columncolor{gray!25}}c}{\textbf{49.79}} & \multicolumn{1}{>{\columncolor{gray!25}}c}{72.34} & \multicolumn{1}{>{\columncolor{gray!25}}c}{43.47} & \multicolumn{1}{>{\columncolor{gray!25}}c}{43.07} \\
				& \multicolumn{1}{>{\columncolor{gray!25}}c}{+FSC (n=2)} &
				\multicolumn {1}{>{\columncolor {gray!25}} c}{88.79} & \multicolumn {1}{>{\columncolor {gray!25}} c}{\textbf{79.99}} & \multicolumn {1}{>{\columncolor {gray!25}} c}{50.00} & \multicolumn {1}{>{\columncolor {gray!25}} c}{60.93} & \multicolumn {1}{>{\columncolor {gray!25}} c}{79.94} & \multicolumn {1}{>{\columncolor {gray!25}} c}{\textbf{81.09}} & \multicolumn {1}{>{\columncolor {gray!25}} c}{\textbf{87.48}} & \multicolumn {1}{>{\columncolor {gray!25}} c}{90.87} & \multicolumn {1}{>{\columncolor {gray!25}} c}{82.18} & \multicolumn {1}{>{\columncolor {gray!25}} c}{76.41} & \multicolumn {1}{>{\columncolor {gray!25}} c}{56.68} & \multicolumn {1}{>{\columncolor {gray!25}} c}{62.01} & \multicolumn {1}{>{\columncolor {gray!25}} c}{70.54} & \multicolumn {1}{>{\columncolor {gray!25}} c}{72.60} & \multicolumn {1}{>{\columncolor {gray!25}} c}{46.01} & \multicolumn {1}{>{\columncolor {gray!25}} c}{\textbf{72.37}} & \multicolumn {1}{>{\columncolor {gray!25}} c}{\textbf{43.96}} & \multicolumn {1}{>{\columncolor {gray!25}} c}{\textbf{43.10}} \\
				\hline
				\multirow{7}{*}{\rotatebox{90}{\parbox{2cm}{\centering RetinaNet}}} 
				& Direct &80.40 & 69.60 & 38.80 & 66.50 & 70.80 & \textbf{69.60} & \textbf{87.10} & 89.50 & 68.00 & 74.60 & 49.30 & 62.80 & 58.40 & \textbf{86.70} & 45.50 & 68.30&41.50&40.79\\

				
				& +KLD & \textbf{88.98} & 76.21 & 41.64 & 64.11 & 78.31 & 63.44 & 77.66 & 90.87 & 80.99 & \textbf{83.02} & 59.02 & 63.66 & 57.50 & 66.51 & 44.13 & 69.07 & 37.84 & 38.89 \\
				& +CSL & 89.33 & 79.67 & 40.83 & 69.95 & 77.71 & 62.08 & 77.46 & 90.87 & 82.87 & 82.03 & 60.07 & 65.27 & 53.58 & 64.03 & 46.62 & 69.49 & 40.42 & 39.69 \\
				& +PSC & 89.41 & 80.66 & 39.06 & 69.08 & 77.61 & 61.63 & 77.21 & 90.86 & 82.52 & 81.76 & \textbf{60.98} & \textbf{66.20} & 57.51 & 64.75 & 48.28 & 69.83 & 40.37 & 40.03 \\
				& +PSCD & 89.32 & \textbf{82.29} & 37.92 & \textbf{71.52} & 78.40 & 66.33 & 78.01 & \textbf{90.89} & 84.21 & 80.63 & 60.22 & 64.73 & 59.69& 68.37 & \textbf{53.85} & 71.09 & 41.17 & 41.25 \\
				
				
				& \multicolumn{1}{>{\columncolor{gray!25}}c}{+FSC (n=1)} & \multicolumn{1}{>{\columncolor{gray!25}}c}{88.18} & \multicolumn{1}{>{\columncolor{gray!25}}c}{81.50} & \multicolumn{1}{>{\columncolor{gray!25}}c}{44.50} & \multicolumn{1}{>{\columncolor{gray!25}}c}{70.01} & \multicolumn{1}{>{\columncolor{gray!25}}c}{\textbf{79.26}} & \multicolumn{1}{>{\columncolor{gray!25}}c}{69.46} & \multicolumn{1}{>{\columncolor{gray!25}}c}{81.91} & \multicolumn{1}{>{\columncolor{gray!25}}c}{90.63} & \multicolumn{1}{>{\columncolor{gray!25}}c}{82.36} & \multicolumn{1}{>{\columncolor{gray!25}}c}{81.51} & \multicolumn{1}{>{\columncolor{gray!25}}c}{58.86} & \multicolumn{1}{>{\columncolor{gray!25}}c}{66.81} & \multicolumn{1}{>{\columncolor{gray!25}}c}{62.38} & \multicolumn{1}{>{\columncolor{gray!25}}c}{68.61} & \multicolumn{1}{>{\columncolor{gray!25}}c}{\textbf{57.09}} & \multicolumn{1}{>{\columncolor{gray!25}}c}{\textbf{72.21}} & \multicolumn{1}{>{\columncolor{gray!25}}c}{\textbf{45.56}} & \multicolumn{1}{>{\columncolor{gray!25}}c}{\textbf{43.45}} \\
				
				& \multicolumn{1}{>{\columncolor{gray!25}}c}{+FSC (n=2)} &
				\multicolumn{1}{>{\columncolor{gray!25}}c}{88.39} & \multicolumn{1}{>{\columncolor{gray!25}}c}{81.90} & \multicolumn{1}{>{\columncolor{gray!25}}c}{\textbf{44.77}} & \multicolumn{1}{>{\columncolor{gray!25}}c}{70.17} & \multicolumn{1}{>{\columncolor{gray!25}}c}{79.11} & \multicolumn{1}{>{\columncolor{gray!25}}c}{69.36} & \multicolumn{1}{>{\columncolor{gray!25}}c}{80.91} & \multicolumn{1}{>{\columncolor{gray!25}}c}{90.76} & \multicolumn{1}{>{\columncolor{gray!25}}c}{\textbf{84.82}} & \multicolumn{1}{>{\columncolor{gray!25}}c}{82.28} & \multicolumn{1}{>{\columncolor{gray!25}}c}{59.14} & \multicolumn{1}{>{\columncolor{gray!25}}c}{65.80} & \multicolumn{1}{>{\columncolor{gray!25}}c}{\textbf{62.63}} & \multicolumn{1}{>{\columncolor{gray!25}}c}{67.55} & \multicolumn{1}{>{\columncolor{gray!25}}c}{51.18} & \multicolumn{1}{>{\columncolor{gray!25}}c}{71.92} & \multicolumn{1}{>{\columncolor{gray!25}}c}{44.99} & \multicolumn{1}{>{\columncolor{gray!25}}c}{42.86} \\

				\hline
				\multirow{7}{*}{\rotatebox{90}{\parbox{2cm}{\centering ${\text{S}^\text{2}}\text{A-Ne}{{\text{t}}}$}}} 
				
				& Direct & 88.89 & 82.53 & 53.37 & 73.62 & 79.40 & 81.01 & 87.62 & 90.90 & 84.54 & 84.57 & 64.14 & 66.38 & 75.12 & 70.97 & 47.58 & 75.38 & 48.48 & 45.58 \\
				
				& +KLD & 88.67 & 82.89 & 53.52 & 74.75 & 78.96 & 82.35 & 87.54 & 90.89 & 84.66 & 85.49 & 68.21 & 66.33 & 75.35 & 69.57 & 54.77 & 76.26 & 48.43 & 45.80 \\
				
				& +CSL & 89.17 & 83.05 & 55.71 & 77.57 & \textbf{79.77} & 82.36 & 88.16 & 90.86 & 86.61 & 85.31 & 67.46 & 63.96 & 76.95 & \textbf{71.83} & 54.85 & 76.91 & 48.61 & 46.26 \\
				
				& +PSC & \textbf{89.32} & 82.93 & 55.62 & \textbf{78.90} & 79.53 & 82.27 & \textbf{88.23} & \textbf{90.90} & 86.93 & \textbf{86.08} & 65.86 & 67.51 & 76.08 & 71.44 & 54.82 & 77.09 & 50.21 & 46.77 \\
				& +PSCD & 88.93 & \textbf{84.46} & 56.59 & 77.16 & 79.30 & \textbf{83.12} & 88.01 & 90.87 & 86.38 & 85.49 & 66.20 & 67.47 & 76.86 & 71.42 & 52.91 & 77.01 & \textbf{50.51} & 46.94 \\
				
				& \multicolumn{1}{>{\columncolor{gray!25}}c}{FSC (n=1)} &
				\multicolumn{1}{>{\columncolor{gray!25}}c}{89.12} & \multicolumn{1}{>{\columncolor{gray!25}}c}{84.42} & \multicolumn{1}{>{\columncolor{gray!25}}c}{\textbf{57.47}} & \multicolumn{1}{>{\columncolor{gray!25}}c}{76.09} & \multicolumn{1}{>{\columncolor{gray!25}}c}{79.48} & \multicolumn{1}{>{\columncolor{gray!25}}c}{83.07} & \multicolumn{1}{>{\columncolor{gray!25}}c}{88.16} & \multicolumn{1}{>{\columncolor{gray!25}}c}{90.87} & \multicolumn{1}{>{\columncolor{gray!25}}c}{\textbf{87.59}} & \multicolumn{1}{>{\columncolor{gray!25}}c}{85.66} & \multicolumn{1}{>{\columncolor{gray!25}}c}{64.69} & \multicolumn{1}{>{\columncolor{gray!25}}c}{68.85} & \multicolumn{1}{>{\columncolor{gray!25}}c}{76.38} & \multicolumn{1}{>{\columncolor{gray!25}}c}{71.61} & \multicolumn{1}{>{\columncolor{gray!25}}c}{55.36} & \multicolumn{1}{>{\columncolor{gray!25}}c}{77.25} & \multicolumn{1}{>{\columncolor{gray!25}}c}{49.60} & \multicolumn{1}{>{\columncolor{gray!25}}c}{\textbf{46.80}} \\
				
				& \multicolumn{1}{>{\columncolor{gray!25}}c}{FSC (n=2)} &
				\multicolumn{1}{>{\columncolor{gray!25}}c}{88.68} & \multicolumn{1}{>{\columncolor{gray!25}}c}{84.04} & \multicolumn{1}{>{\columncolor{gray!25}}c}{56.59} & \multicolumn{1}{>{\columncolor{gray!25}}c}{78.10} & \multicolumn{1}{>{\columncolor{gray!25}}c}{79.29} & \multicolumn{1}{>{\columncolor{gray!25}}c}{82.86} & \multicolumn{1}{>{\columncolor{gray!25}}c}{88.07} & \multicolumn{1}{>{\columncolor{gray!25}}c}{90.89} & \multicolumn{1}{>{\columncolor{gray!25}}c}{87.22} & \multicolumn{1}{>{\columncolor{gray!25}}c}{85.98} & \multicolumn{1}{>{\columncolor{gray!25}}c}{\textbf{68.27}} & \multicolumn{1}{>{\columncolor{gray!25}}c}{\textbf{69.95}} & \multicolumn{1}{>{\columncolor{gray!25}}c}{\textbf{77.11}} & \multicolumn{1}{>{\columncolor{gray!25}}c}{70.46} & \multicolumn{1}{>{\columncolor{gray!25}}c}{\textbf{55.69}} & \multicolumn{1}{>{\columncolor{gray!25}}c}{\textbf{\textbf{77.55}}} & \multicolumn{1}{>{\columncolor{gray!25}}c}{50.14} & \multicolumn{1}{>{\columncolor{gray!25}}c}{46.62} \\
				
				\hline
				%
				%
			\end{tabular}
		}
	\end{table*}
	
	As shown in the Fig. \ref{component fluctuations}, for the single frequency where periodic correction is not required, the component fluctuations of FSC and PSC are negligible.
	To provide deeper insights into how the proposed manifold constraint mitigates decoding instability, we visualize the continuous tracking of raw output components alongside their corresponding decoded angles. Since the decoding process strictly relies on the $\arctan2$ function, the final angular stability is inherently dependent on the structural integrity (i.e., the modulus) of these intermediate components.
	As illustrated in Fig. \ref{Vis_manifold}\textcolor{red}{(a)}, the unconstrained PSCD baseline struggles to fit the theoretical sinusoidal curves, exhibiting noticeable component drift and modulus decay. While the non-orthogonal components fail to maintain a stable Cartesian amplitude, the network becomes highly vulnerable to underlying predictive noise. Once the feature modulus drops near the origin, the $\arctan2$ decoding phase becomes extremely chaotic. Consequently, this feature-level instability directly triggers substantial phase-wrapping singularities, as evidenced by the severe cycle errors shown in Fig. \ref{Vis_manifold}\textcolor{red}{(c)}.
	
	In contrast, Fig. \ref{Vis_manifold}\textcolor{red}{(b)} demonstrates the profound regulatory effect of integrating explicit geometric regularization. By enforcing the sum-of-squares manifold constraint, the network is mathematically guided to lock its predicted components closer to the theoretical unit circle. As seen in Fig. \ref{Vis_manifold}\textcolor{red}{(b)}, the constrained components exhibit much smoother trajectories that more tightly align with the ground-truth curves, effectively alleviating the severity of modulus collapse. 
	Benefiting from this improved feature representation, the subsequent decoding process becomes significantly more robust. As visualized in Fig. \ref{Vis_manifold}\textcolor{red}{(d)}, while a minor fraction of singular points may still occasionally occur under extreme noise, the overall cyclic ambiguity and boundary mutations are drastically reduced compared to the baseline. This continuous, high-fidelity angular tracking explicitly demonstrates that maintaining the feature modulus is a critical mechanism for alleviating cycle errors.

\textbf{Loss Weights.} As shown in Tab. \ref{tab:ablation_angle_loss1}, we perform ablation experiments with different loss settings. Specifically, the detection performance of the FSC improved significantly over the baseline (with $\lambda_{2}=0$), demonstrating its effectiveness. As mentioned earlier, using single-frequency FSC (n=1) is more suitable for rectangular objects than using the dual-frequency FSC. Therefore, for the HRSC-2016 dataset, we set \(\lambda_1\) to 0.7 and \(\lambda_2\) to 0.6. For DIOR-R and DOTA-v1.0 datasets, the default settings for \(\lambda_1\) and \(\lambda_2\) are 1.0 and 0.2, respectively.
	
		\begin{figure*}[t]
		\centering
		\subfloat[Baseline: FCOS (r-50). \label{ship_base}]{%
			\includegraphics[width=0.25\linewidth]{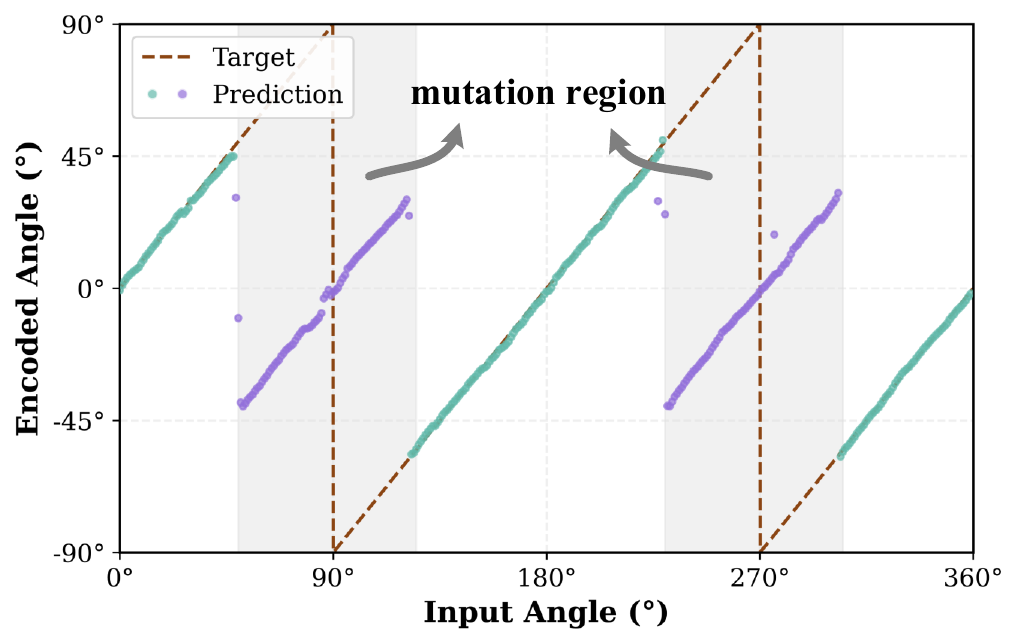}%
		}
		\hfil
		\subfloat[PSCD. \label{rr_psc_ship}]{%
			\includegraphics[width=0.25\linewidth]{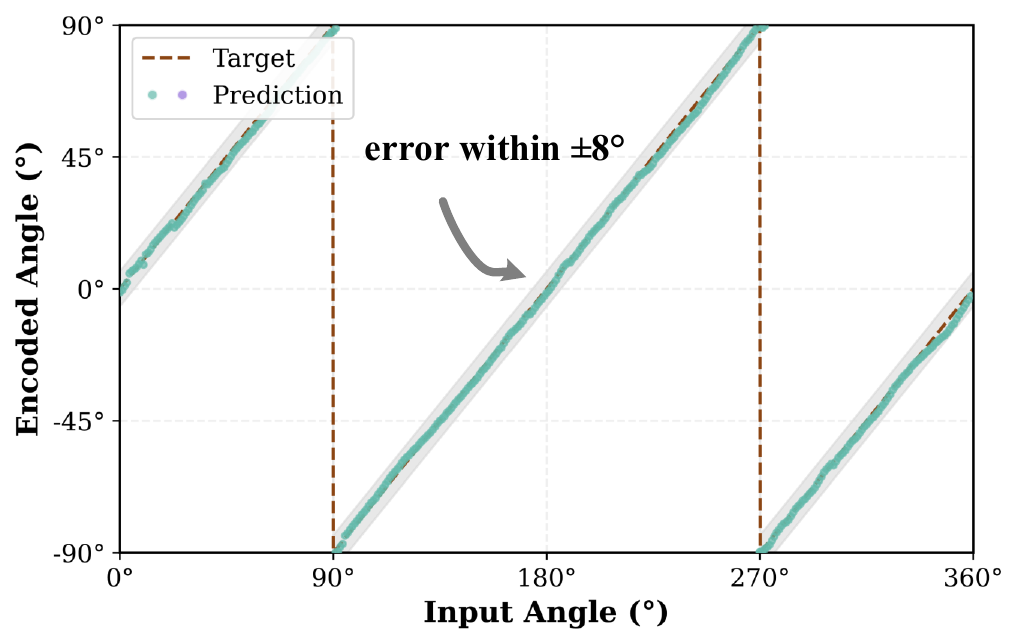}%
		}
		\hfil
		\subfloat[FSC (n=1). \label{ship_fsc1}]{%
			\includegraphics[width=0.25\linewidth]{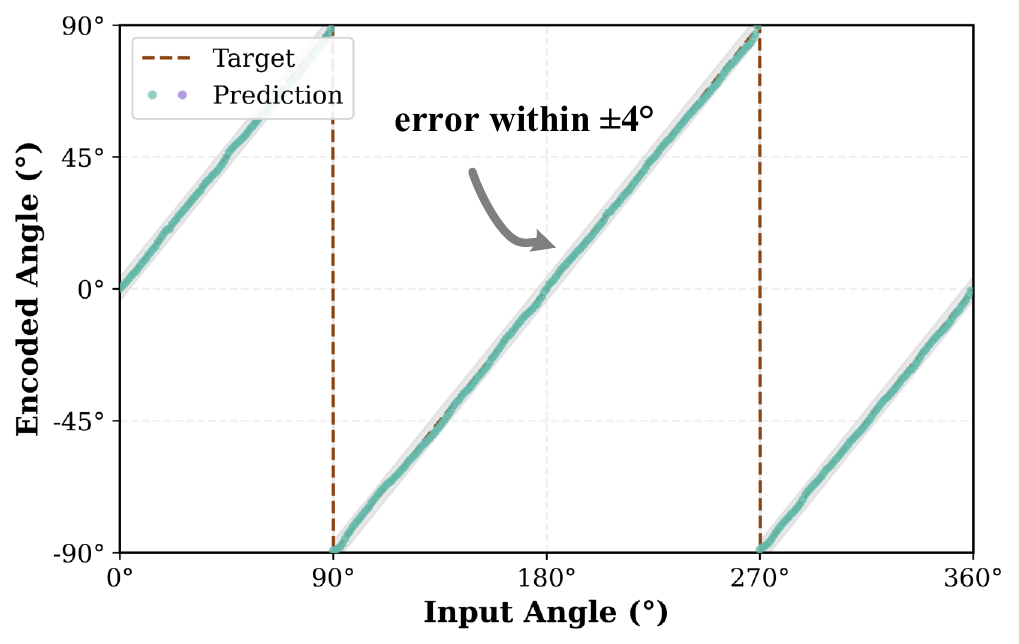}%
		}
		\hfil
		\subfloat[FSC (n=2). \label{ship_fsc2}]{%
			\includegraphics[width=0.25\linewidth]{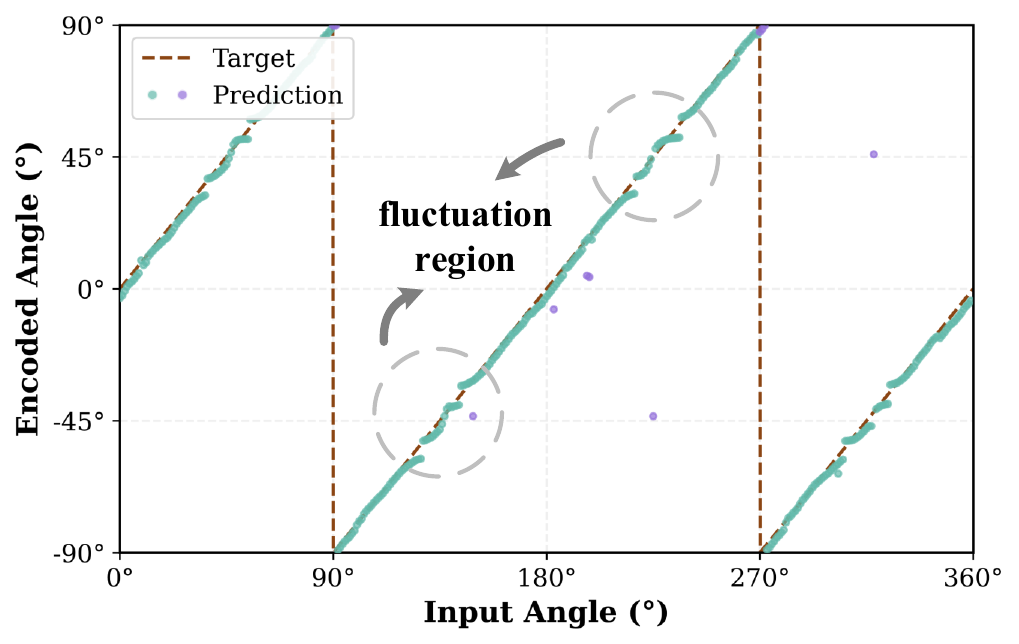}%
		}
		\hfil
		\subfloat[Baseline: ${\text{S}^\text{2}}\text{A-Ne}{{\text{t}}}$ (r-50). \label{s2_base}]{%
			\includegraphics[width=0.25\linewidth]{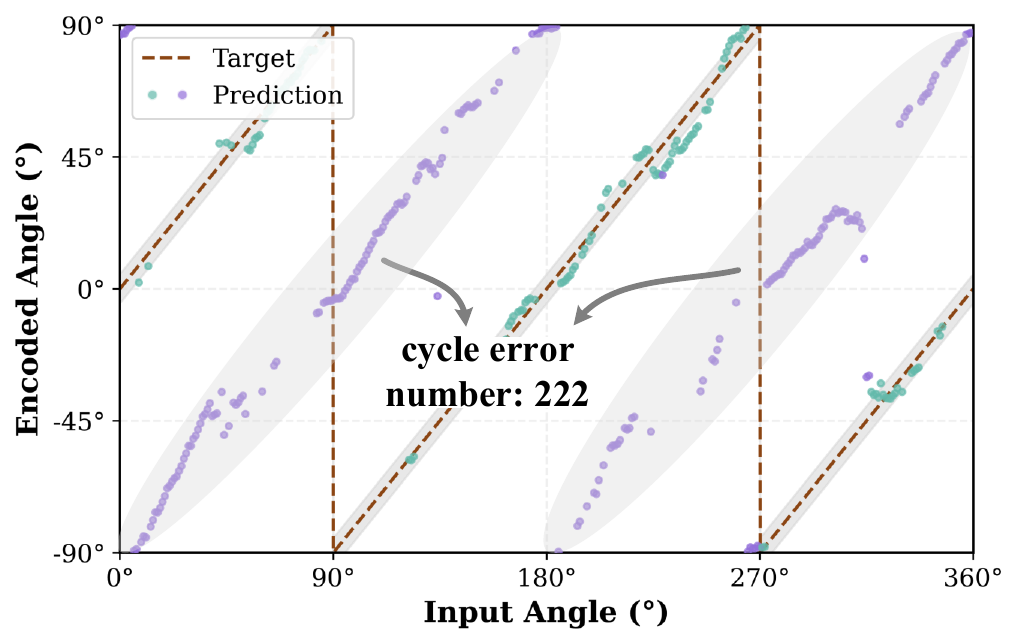}%
		}
		\hfil
		\subfloat[PSCD. \label{s2_dpsc}]{%
			\includegraphics[width=0.25\linewidth]{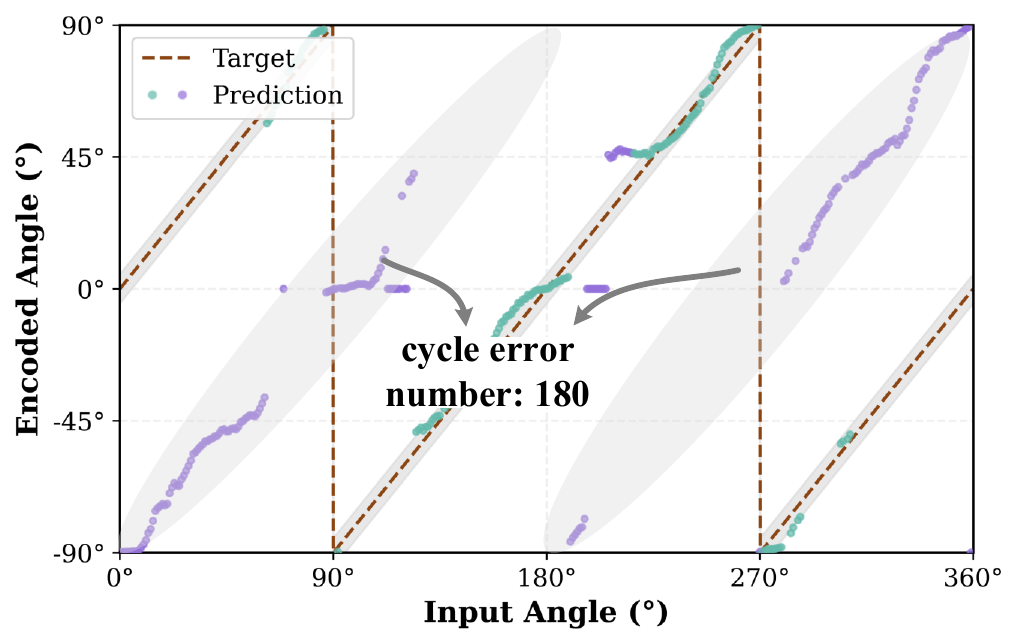}%
		}
		\hfil
		\subfloat[FSC (n=1). \label{s2_fsc1}]{%
			\includegraphics[width=0.25\linewidth]{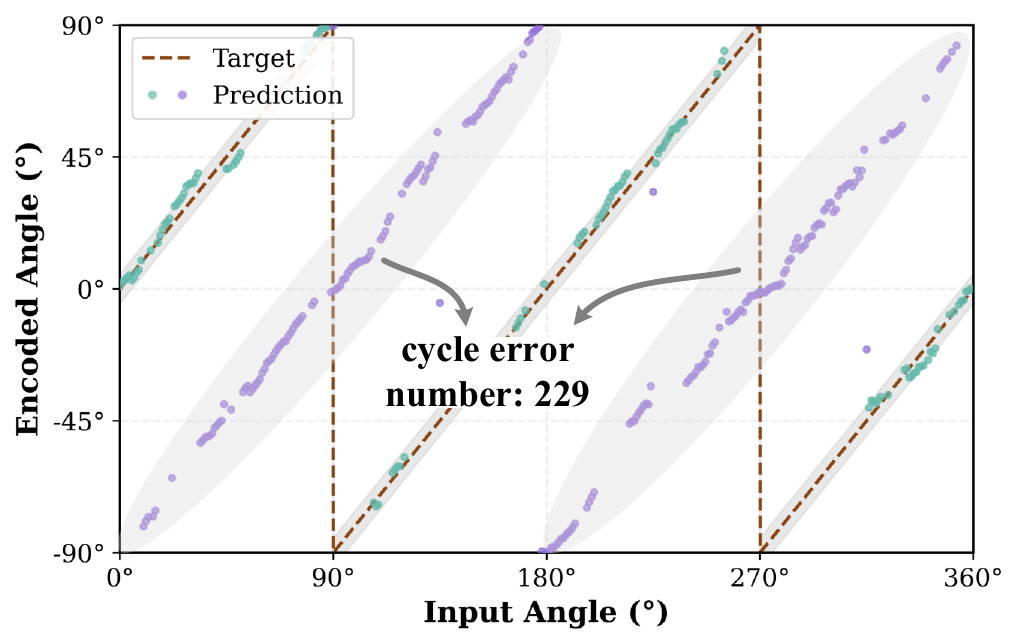}%
		}
		\hfil
		\subfloat[FSC (n=2). \label{s2_fsc2}]{%
			\includegraphics[width=0.25\linewidth]{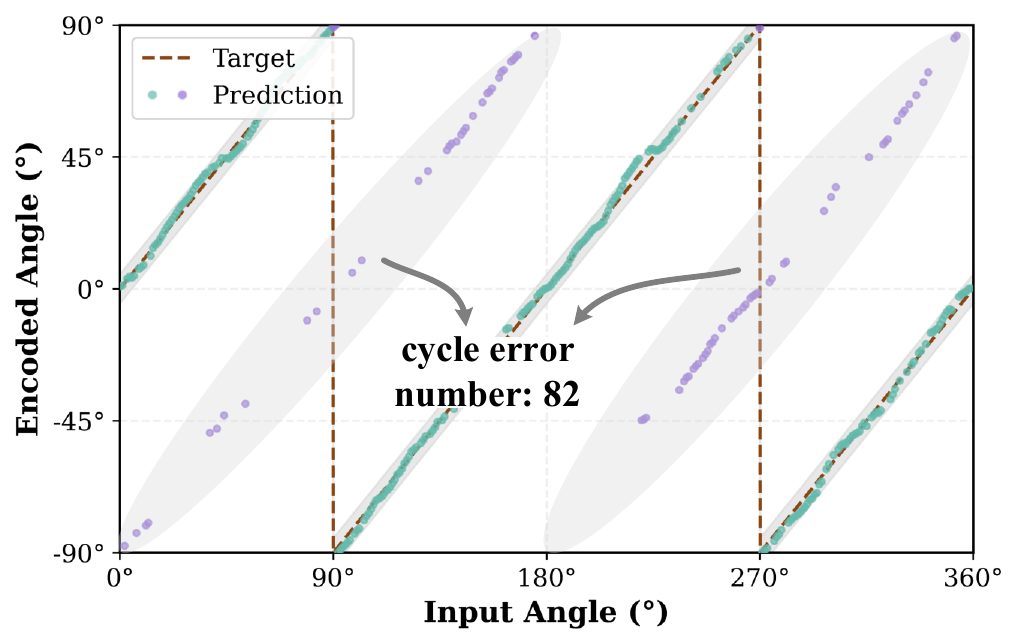}%
		}
		\caption{Illustration of the angle results with different angle coder settings. Green dots indicate correct detection within the allowable range, while purple dots represent erroneous detection.}
		\label{fsc}
	\end{figure*}

	\begin{table*}[t]
		\centering
		\caption{Comparisons with state-of-the-art detectors on \textbf{DOTA-v1.0} dataset. $\ddagger$ indicates multi-scale training/testing.}
		\resizebox{\linewidth}{!}{
			\begin{tabular}{c|c|ccccccccccccccc|c}
				\toprule
				\textbf{Method} & \textbf{Backbone} & \textbf{PL} & \textbf{BD} & \textbf{BR} & \textbf{GTF} & \textbf{SV} & \textbf{LV} & \textbf{SH} & \textbf{TC} & \textbf{BC} & \textbf{ST} & \textbf{SBF} & \textbf{RA} & \textbf{HA} & \textbf{SP} & \textbf{HC} & \textbf{AP$_{50}$} \\
				\midrule 
				PIoU \cite{chen2020piou} & R-50 & 80.90 & 69.70 & 24.10 & 60.20 & 38.30 & 64.40 & 64.80 & \textbf{90.90} & 77.20 & 70.40 & 46.50 & 37.10 & 57.10 & 61.90 & 64.00 & 60.50 \\
				Gliding Vertex$^\ddagger$ \cite{xu2020gliding} & R-50 & 89.64 & 85.00 & 52.26 & 77.34 & 73.01 & 73.14 & 86.82 & 90.74 & 79.02 & 86.81 & 59.55 & 70.91 & 72.94 & 70.86 & 57.32 & 75.02 \\
				CSL$^\ddagger$ \cite{yang2020arbitrary} & R-50 & \textbf{90.25} & 85.53 & 54.64 & 75.31 & 70.44 & 73.51 & 77.62 & 90.84 & 86.15 & 86.69 & 69.60 & 68.04 & 73.83 & 71.10 & 68.93 & 76.17 \\
				R$^3$Det$^\ddagger$ \cite{yang2021r3det} & R-50 & 89.80 & 83.77 & 48.11 & 66.77 & 78.76 & 83.27 & 87.84 & 90.82 & 85.38 & 85.51 & 65.67 & 62.68 & 67.53 & 78.56 & 72.62 & 76.47 \\
				GWD$^\ddagger$ \cite{yang2021rethinking} & R-50 & 86.96 & 83.88 & 54.36 & 77.53 & 74.41 & 68.48 & 80.34 & 86.62 & 83.41 & 85.55 & 73.47 & 67.77 & 72.57 & 5.76 & 73.40 & 76.30 \\
				SCRDet++$^\ddagger$ \cite{yang2022scrdet++} & R-50 & 90.05 & 84.39 & 55.44 & 73.99 & 77.54 & 71.11 & 86.05 & 90.67 & 87.32 & 87.08 & 69.62 & 68.90 & 73.74 & 71.29 & 65.08 & 76.81 \\
				KFIoU$^\ddagger$ \cite{yang2022kfiou} & R-50 & 89.46 & 85.72 & 54.94 & 80.37 & 77.16 & 69.23 & 80.90 & 90.79 & \textbf{87.79} & 86.13 & \textbf{73.32} & 68.11 & 75.23 & 71.61 & 69.49 & 77.35 \\
				DCL$^\ddagger$ \cite{yang2021dense} & R-50 & 89.26 & 83.60 & 53.54 & 72.76 & 79.04 & 82.56 & 87.31 & 90.67 & 86.59 & 86.98 & 67.49 & 66.88 & 73.29 & 70.56 & 69.99 & 77.37 \\
				PSC$^\ddagger$ \cite{yu2023phase} & R-50 & 89.86 & 86.02 & 54.94 & 62.02 & \textbf{81.90} & \textbf{85.48} & 88.39 & 90.73 & 86.90 & \textbf{88.82} & 63.94 & 69.19 & 76.84 & \textbf{82.75} & 63.24 & 78.07 \\
				KLD$^\ddagger$ \cite{yang2021learning} & R-50 & 88.91 & 85.23 & 53.64 & 81.23 & 78.20 & 76.99 & 84.58 & 89.50 & 86.84 & 86.38 & 71.69 & 68.06 & 75.95 & 72.23 & \textbf{75.42} & 78.32 \\
				ACM$^\ddagger$ \cite{xu2024rethinking} & R-50 & 89.84 & 85.50 & 53.84 & 74.78 & 80.77 & 82.81 & 88.92 & 90.82 & 87.18 & 86.53 & 64.09 & 66.27 & 77.51 & 79.62 & 69.57 & 78.53 \\
				S$^2$A-Net \cite{han2021align}& R-50 & 89.11 & 82.84 & 48.37 & 71.11 & 78.11 & 78.39 & 87.25 & 90.83 & 84.90 & 85.64 & 60.36 & 62.60 & 65.26 & 69.13 & 57.94 & 74.12 \\
				S$^2$A-Net$^\ddagger$ \cite{han2021align} & R-50 & 89.28 & 84.11 & 56.95 & 79.21 & 80.18 & 82.93 & \textbf{89.21} & 90.86 & 84.66 & 87.61 & 71.66 & 68.23 & 78.58 & 78.20 & 65.55 & 79.15 \\
				GSDet$^\ddagger$ \cite{dinggsdet} & R-50 & 88.04 & 85.55 & 57.22 & 79.44 & 81.28 & 84.77 & 88.73 & 90.82 & 87.11 & 87.32 & 68.64 & 68.54 & \textbf{79.57} & 80.81 & 72.79 & 80.04 \\
				GSDet \cite{dinggsdet} & R-50 & 88.65 & 81.31 & 51.11 & 73.48 & 78.59 & 82.76 & 88.17 & 90.83 & 84.04 & 81.35 & 59.68 & 62.27 & 74.68 & 69.91 & 64.89 & 75.44 \\
				GauCho$^\ddagger$ \cite{marques2025gaucho} & R-50 & 88.96 & 81.01 & 57.39 & 72.21 & 82.40 & 85.41 & 88.51 & 90.85 & 85.42 & 86.40 & 66.42 & 70.19 & 76.10 & 80.42 & 71.00 & 78.85 \\
				FAAFusion$^\ddagger$ \cite{gu2026fourier} & R-50 & 89.81 & 83.27 & 52.93 & 77.06 & 79.57 & 84.90 & 88.31 & 90.89 & 87.10 & 86.43 & 66.43 & 68.04 & 75.54 & 79.03 & 71.48 & 78.72 \\
				\hline
%
				
					\rowcolor{gray!25}	
				S$^2$A-Net (Ours)& R-50 & 88.68 & 84.04 & 56.59 & 78.10 & 79.29 & 82.86 & 88.07 & 90.89 & 87.22 & 85.98 & 68.27 & 69.95 & 77.11 & 70.46 & 55.69 & 77.55 \\
				
				\rowcolor{gray!25}	
				S$^2$A-Net$^\ddagger$ (Ours) & R-50 & 89.32 & \textbf{85.70} & \textbf{59.30} & \textbf{81.11} & 80.22 & 84.91 & 88.88 & 90.80 & 85.89 & 87.49 & 72.38 & \textbf{69.77} & 78.46 & 81.30 & 71.53 & \textbf{80.47} \\
				\bottomrule
			\end{tabular}
		}
		\label{tab:dota_performance}
	\end{table*}
\subsection{Comparisons with State-of-the-arts}

\textbf{Comparisons on DOTA-v1.0 Dataset.} Tab. \ref{tab:dota_ap50} compares with state-of-the-art OBB/angle coders. Experiments demonstrate that our FSC achieves highly competitive performance on the large-scale DOTA-v1.0 dataset.
On the RetinaNet baseline, FSC (n=1) and PSCD yield nearly identical AP$_{50}$. Crucially, FSC achieves 4.39\% improvement in AP$_{75}$. 
Notably, CSL's performance remains largely invariant to rotation data augmentation. In contrast, FSC improves its AP$_{75}$ by 1.50\% and 1.76\% in two frequency settings in FCOS, respectively. By utilizing its unique encoding mechanism (i.e., the DC component), FSC can fit the angular distribution more accurately.

To comprehensively demonstrate the superiority of FSC, Fig. \ref{vis_dota} visualizes the detection results of CSL, PSCD, and our FSC (n=2) across diverse and complex large-scale scenes in the DOTA-v1.0 dataset. 
As observed in the first row of Fig. \ref{vis_dota}, CSL struggles to maintain stable orientation predictions, primarily due to the inherent quantization errors of its discrete classification mechanism. While PSCD theoretically addresses the boundary discontinuity and square-like object problems, the absence of a manifold constraint in its design leads to substantial angular deviations during decoding. In contrast, our FSC (n=2) demonstrates exceptional robustness and precise alignment capabilities across all complex scenarios.
Furthermore, this visual comparison reveals a critical insight regarding the nonlinear impact of angular deviation on the Intersection over Union (IoU) metric. For objects with large aspect ratios (e.g., ship, bridge, and harbor), even a minute angular offset can cause a dramatic drop in IoU, leading to severe false positives or missed detections. This extreme angular sensitivity corroborates the findings of Liu \textit{et al.} \cite{hou2022refined}, who similarly highlighted the severe degradation of bounding box alignment for objects with extreme aspect ratios. Conversely, for square-like objects (e.g., airplane, storage tank, and baseball diamond), minor angular deviations have a relatively negligible impact on IoU, provided the center points are accurately aligned. This intrinsic geometric property has been largely overlooked by many classic works (e.g., CSL \cite{yang2020arbitrary} and GWD \cite{yang2021rethinking}). By robustly decoding the orientation, FSC ensures highly accurate alignment regardless of the object's aspect ratio.

As shown in Fig. \ref{fsc}, we visualize the aforementioned image series, where each image contains a single object. Angle predictions of PSCD and FSC are compared on two baselines, FCOS \cite{tian2019fcos} and ${\text{S}^\text{2}}\text{A-Net}$ \cite{han2021align}. In Fig. \ref{fsc}\textcolor{red}{(a)}, green dots indicate correct detections (i.e., those within the allowable error range), while purple dots denote incorrect ones. The mutation region refers to the model's lazy region, in which detection results meet a specific IoU threshold but lack angular accuracy. 
For example, the 45° object in Fig. \ref{fsc}\textcolor{red}{(a)} is predicted as -45°, yet similar bounding boxes still contribute to AP$_{50}$. 
As illustrated in Fig. \ref{fsc}\textcolor{red}{(b)}, PSCD employs dual-frequency components for decoding, restricting the error to $\pm8^\circ$. For Fig. \ref{fsc}\textcolor{red}{(c)}, FSC (n=1) uses only single-frequency decoding, limiting the error to within $\pm4^\circ$. Under the stronger constraint of $n=2$, however, the prediction results exhibit fluctuation regions, as shown in Fig. \ref{fsc}\textcolor{red}{(d)}. We attribute this phenomenon to the fact that rectangular objects only require single-frequency decoding, whereas redundant frequency information subjects the model to noise interference.   
When extended to square objects (second row of Fig. \ref{fsc}), their angle predictions show more severe fluctuations. Our FSC (n=2) not only delivers more stable detection performance but also minimizes cycle errors.

Tab. \ref{tab:dota_performance} provides a detailed comparison of our method with state-of-the-art detectors on the DOTA-v1.0 dataset. 
Note that competing methods vary in their core functional components (e.g., OBB/angle coders, loss functions) and network architectures.
Despite these discrepancies, our FSC implemented with ${\text{S}^\text{2}}\text{A-Ne}{{\text{t}}}$ (r-50) achieves a competitive AP$_{50}$ of 80.47 under relatively fair experimental conditions.

\begin{figure}[t]
	\centering{\includegraphics[width=\columnwidth]{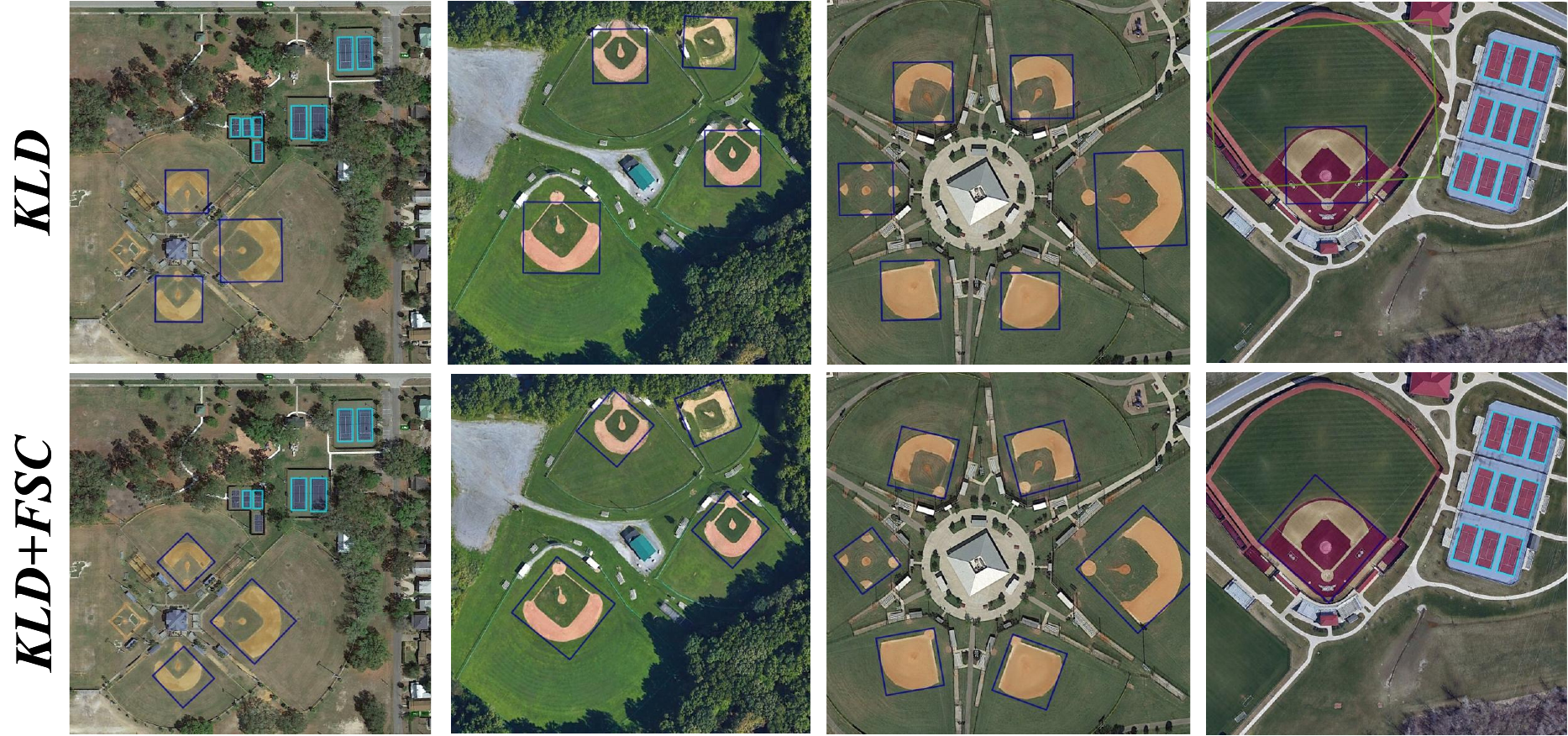}}
	\caption{Visual comparison of KLD \cite{yang2021learning} and $\text{KLD}^{+}$ (ours) for detecting square-like objects..}
	\label{kld_fsc2}
\end{figure}

\begin{table}[t]
	\centering
	\caption{Comparison with Gaussian-based methods w/ and w/o FSC in terms of in AP $\text{(\%)}$ and parameters $\text{(M)}$ on the DIOR-R dataset.}
	\label{tab:dior}
	\resizebox{\linewidth}{!}{
		\begin{tabular}{l|cc|cc}
			\hline
			Method & $\text{AP}_{50}$ & $\text{AP}_{75} $& Params\\
			
			\hline
			GWD & 54.80 & 34.00&36.52\\
			\rowcolor{gray!25} $\text{GWD}^{+}$ & 58.90 \textcolor{mygreen}{(+4.10)} &  39.00 \textcolor{mygreen}{(+5.00)} &36.60 \textcolor{red}{(+0.08)}\\
			\hline
			KLD  & 55.60 &34.40 &36.52\\
			\rowcolor{gray!25} $\text{KLD}^{+}$ & 59.80 \textcolor{mygreen}{(+4.20)} &  37.70 \textcolor{mygreen}{(+3.30)} &36.60 \textcolor{red}{(+0.08)}\\
			\hline
	\end{tabular}}
\end{table}

\textbf{Comparisons on DIOR-R Dataset.} To explore the robust performance of FSC, experiments are conducted on the DIOR-R dataset, as shown in Tab. \ref{tab:dior}. FSC can be seamlessly integrated with Gaussian-based methods (e.g., GWD and KLD) by introducing it as an explicit angle regression branch alongside the Gaussian bounding box loss. In this decoupled paradigm, the Gaussian loss governs the robust optimization of spatial parameters ($x, y, w, h$), while FSC explicitly supervises the orientation $\theta$ through its manifold-constrained Fourier components. 

This combination delivers significant performance improvements with a negligible increase in parameters. Specifically, it enhances $\text{AP}_{50}$ by $3.00\%$ and $3.70\%$, and $\text{AP}_{75}$ by $2.70\%$ and $3.30\%$, respectively. We observe that the increments in the $\text{AP}_{75}$ and $\text{AP}_{50}$ metrics tend to be close, suggesting optimized high-precision performance in angle prediction. 

Fundamentally, FSC mitigates the inherent limitations of Gaussian-based methods in handling square-like objects. Fig. \ref{kld_fsc2} visually contrasts the results of pure KLD \cite{yang2021learning} with our combined KLD+FSC paradigm. Since mathematically identical Gaussian distributions can represent distinct physical orientations for square-like objects, KLD is significantly hindered by Cyclic Ambiguity (CA). By enforcing explicit angular constraints via the dual-frequency orthogonal basis, FSC effectively eliminates this ambiguity, allowing the combined model to accurately capture the true orientation of such objects.

	\begin{table}[t]
	\centering
	\scriptsize
	\tiny
	\caption{Comparison with state-of-the-art OBB/angle coders on the \textbf{HRSC-2016} dataset. \textcolor{red}{Red} and \textcolor{blue}{blue}: top two performances.}
	\label{tab:angle_encoding_comparison_empty}
	\resizebox{\linewidth}{!}{
		\begin{tabular}{l|c|c|ccc}
			\hline
			\multicolumn{2}{c|}{Method} & Length&AP$_{50}$ & AP$_{75}$ &
			AP \\
			
			\hline
			\hline
			\multirow{6}{*}{\rotatebox[origin=c]{90}{FCOS}} & Direct &1& 89.10 & 76.40 &63.06   \\
			& KLD &-&89.20  &66.40  & 58.29 \\
			& CSL &45& 89.40 & 62.50 & 56.27 \\
			& ACM &2& 90.00 & \textcolor{blue}{78.80} & \textcolor{blue}{67.41} \\
			& PSC &3&\textcolor{blue}{90.10}  & 78.00 & 64.93 \\
			& PSCD &6& \textcolor{blue}{90.10} & 78.70 &66.42  \\
			& \multicolumn{1}{>{\columncolor{gray!25}}c|}{FSC (n=1)} & \multicolumn{1}{>{\columncolor{gray!25}}c|}{3} & \multicolumn{1}{>{\columncolor{gray!25}}c}{\textcolor{red}{90.30}} & \multicolumn{1}{>{\columncolor{gray!25}}c}{\textcolor{red}{79.00}} & \multicolumn{1}{>{\columncolor{gray!25}}c}{\textcolor{red}{67.52}} \\
			
			\hline \hline
			\multirow{6}{*}{\rotatebox[origin=c]{90}{RetinaNet}} & Direct &1& 83.90 & 51.60  & 50.41 \\
			& KLD &-& 84.40 & 51.40 &  48.39\\
			& CSL &45&84.20 & 30.20 & 40.62 \\
			& ACM &2&85.20  & \textcolor{blue}{58.10} & \textcolor{blue}{51.31} \\
			& PSC &3&\textcolor{blue}{85.80}  & 50.70 & 50.32 \\
			& PSCD &6&85.70  & 49.40 & 48.91 \\
			& \multicolumn{1}{>{\columncolor{gray!25}}c|}{FSC (n=1)} & \multicolumn{1}{>{\columncolor{gray!25}}c|}{3} & \multicolumn{1}{>{\columncolor{gray!25}}c}{\textcolor{red}{86.40}} & \multicolumn{1}{>{\columncolor{gray!25}}c}{\textcolor{red}{58.50}} & \multicolumn{1}{>{\columncolor{gray!25}}c}{\textcolor{red}{53.04}} \\
			\hline
	\end{tabular}}
\end{table}

\textbf{Comparisons on HRSC-2016 Dataset.} As shown in Tab. \ref{tab:angle_encoding_comparison_empty}, comparative experiments with various state-of-the-art coders are conducted on two baselines, FCOS \cite{tian2019fcos} and RetinaNet \cite{lin2017focal}.
Specifically, CSL improves both AP$_{50}$ by 0.30\%, but AP$_{75}$ significantly decreases by 13.90\% and 21.40\%, respectively. Although CSL strives to design smooth angle labels, its discreteness and encoding length hinder its high-precision performance. Unlike CSL, both PSC and ACM achieve noticeable improvements in both AP$_{50}$ and AP$_{75}$ with continuous encoding. Under the RetinaNet baseline, PSCD and PSC achieve similar AP$_{50}$ results, but suffer a slight decrease in AP$_{75}$. Considering this relationship between encoding length and prediction fluctuation, our FSC demonstrates the best performance across the board.

\subsection{Limitations and Future Work}
\label{Limitations}
While FSC effectively mitigates the ABD and CA problems, it faces an inherent frequency-noise trade-off where higher harmonic orders become increasingly sensitive to neural predictive noise. Mathematically, the rapid oscillation of high-frequency components causes minor regression deviations to be severely amplified during the inverse $\arctan2$ decoding, triggering angular instability. Consequently, this study exclusively utilizes low-frequency components to achieve an optimal balance between resolving topological ambiguity and maintaining robust noise immunity. Future work will explore stabilizing higher-order frequencies via frequency-adaptive filtering or dynamic spectral weighting to further enhance geometric modeling precision.

\section{Conclusion}
\label{Conclusion}

In this paper, we demonstrated that state-of-the-art continuous angle coders are fundamentally limited by severe decoding instability, which stems from unconstrained non-orthogonal designs and feature modulus collapse. To overcome this topological flaw, we proposed the Fourier Series Coder (FSC). By mapping angles into a minimal orthogonal Fourier basis and explicitly enforcing a geometric manifold constraint, FSC inherently prevents modulus decay and ensures structural robustness. Extensive experiments confirm that this mathematical regularization significantly mitigates both the Angle Boundary Discontinuity (ABD) and Cyclic Ambiguity (CA) problems. Ultimately, FSC translates this underlying decoding stability into superior high-precision detection performance, offering a highly reliable paradigm for orientation regression.

%

\bibliography{main.bib}
\bibliographystyle{IEEEtran}

\end{document}